\documentclass[10pt,twocolumn,letterpaper]{article}

\usepackage{style/iccv}
\usepackage{adjustbox}
\usepackage{times}
\usepackage{epsfig}
\usepackage{graphicx}

\usepackage{graphicx}
\usepackage{epsfig}
\usepackage{tikz}
\usetikzlibrary{positioning}
\usetikzlibrary{shapes}
\usetikzlibrary{fit}
\usetikzlibrary{calc}

\usepackage{mathtools,amsfonts,amsmath,amssymb,amsthm}
\theoremstyle{plain}
\newtheorem{theorem}{Theorem}
\newtheorem{proposition}{Proposition}
\newtheorem{lemma}{Lemma}

\usepackage{url}
\usepackage{enumitem}
\usepackage{listings}
\usepackage[ruled,vlined]{algorithm2e}

\usepackage{xcolor}
\definecolor{codegreen}{rgb}{0,0.6,0}
\definecolor{codegray}{rgb}{0.5,0.5,0.5}
\definecolor{codeblack}{rgb}{0.,0.,0.}
\definecolor{codepurple}{rgb}{0.58,0,0.82}
\definecolor{backcolour}{rgb}{0.95,0.95,0.92}

\lstdefinestyle{mystyle}{
    backgroundcolor=\color{backcolour},   
    commentstyle=\color{codegreen},
    keywordstyle=\color{codeblack},
    numberstyle=\tiny\color{codegray},
    stringstyle=\color{codepurple},
    basicstyle=\ttfamily\footnotesize,
    breakatwhitespace=false,         
    breaklines=true,                 
    captionpos=b,                    
    keepspaces=true,                 
    numbers=left,                    
    numbersep=5pt,                  
    showspaces=false,                
    showstringspaces=false,
    showtabs=false,                  
    tabsize=2,
    aboveskip=0pt,
    belowskip=-3pt
}
\usepackage[pagebackref=true,breaklinks=true,colorlinks,bookmarks=false]{hyperref}

\usepackage[capitalize]{cleveref}
\crefname{lemma}{lemma}{lemmas}
\Crefname{lemma}{Lemma}{Lemmas}
\crefname{thm}{theorem}{theorems}
\Crefname{corollary}{Corollary}{Corollaries}

\usepackage{style/math}

\usepackage[accsupp]{axessibility}

\iccvfinalcopy

\def\iccvPaperID{12534} 

\ificcvfinal\fi

\newcommand\blfootnote[1]{%
  \begingroup
  \renewcommand\thefootnote{}\footnote{#1}%
  \addtocounter{footnote}{-1}%
  \endgroup
}

\begin{document}

\title{Active Self-Supervised Learning:\\A Few Low-Cost Relationships Are All You Need}

\author{
Vivien Cabannes*\\
Meta AI
\and
Leon Bottou\\
Meta AI
\and
Yann Lecun\\
Meta AI
\and
Randall Balestriero*\\
Meta AI
}

\maketitle
\ificcvfinal\thispagestyle{empty}\fi

\begin{abstract}
Self-Supervised Learning (SSL) has emerged as the solution of choice to learn transferable representations from unlabeled data.  However, SSL requires to build samples that are known to be semantically akin, i.e. positive views.
Requiring such knowledge is the main limitation of SSL and is often tackled by ad-hoc strategies e.g. applying known data-augmentations to the same input.
In this work, we formalize and generalize this principle through \underline{Positive Active Learning} (PAL) where an oracle queries semantic relationships between samples.
PAL achieves three main objectives.
First, it unveils a theoretically grounded learning framework beyond SSL, based on similarity graphs, that can be extended to tackle supervised and semi-supervised learning depending on the employed oracle.
Second, it provides a consistent algorithm to embed a priori knowledge, e.g. some observed labels, into any SSL losses without any change in the training pipeline.
Third, it provides a proper active learning framework yielding low-cost solutions to annotate datasets, arguably bringing the gap between theory and practice of active learning that is based on simple-to-answer-by-non-experts queries of semantic relationships between inputs.
\end{abstract}

\blfootnote{*Equal contribution.}

\begin{figure*}[t]
    \centering
\begin{adjustbox}{width=.95\textwidth}
\begin{tikzpicture}
\node[inner sep=0pt]  (title) at (0,-.25) {\textbf{Active Learning}};

\node[inner sep=0pt,below=0.5cm of title] (desc)  {
\parbox{6.2cm}{\small\em Given an input, return its label e.g. for Imagenet: 
\texttt{\scriptsize\\- Tinca tinca\\- Carassius auratus\\- Carcharodon carcharias\\- \dots}}};

\node[inner sep=0pt, below left=2.5cm and -0.7cm of title] (q1title) {$\texttt{query}_{\texttt{1}}$};
\node[inner sep=0pt,below=0.1cm of q1title] (q1){\includegraphics[width=2cm,height=2cm]{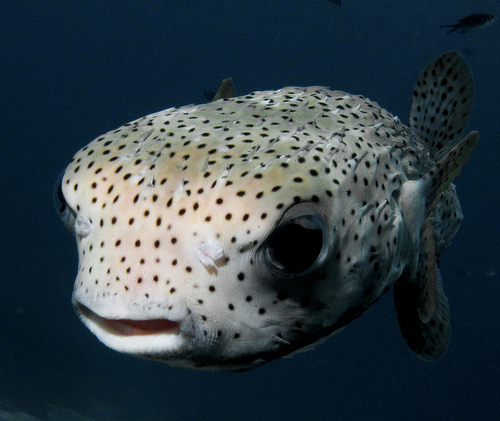}};

\node[inner sep=0pt,below right=2.5cm and -0.7cm of title] (q2title)  {$\texttt{query}_{\texttt{2}}$};
\node[inner sep=0pt,below=.1cm of q2title] (q2) {\includegraphics[width=2cm,height=2cm]{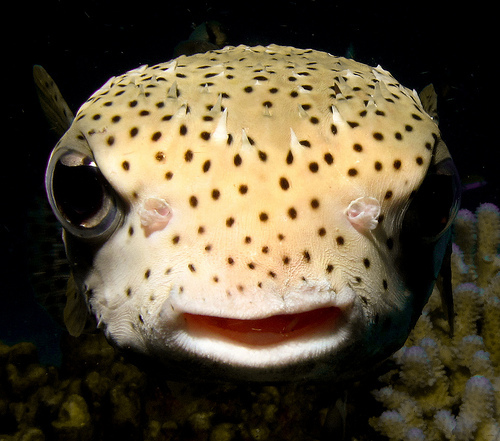}};

\node[inner sep=0pt,below=0.3cm of q1] (option1)  {\parbox{2cm}{\centering$\texttt{response}_{\texttt{1}}$\\ aligator}};

\node[inner sep=0pt,below=0.3cm of q2] (option2)  {\parbox{2cm}{\centering$\texttt{response}_{\texttt{2}}$\\aligator}};

\node[inner sep=0pt,below=6cm of title,color=red] (comment)  {\parbox{7cm}{\centering
Expansive oracle (expert knowledge)}};
    
\draw[black,thick,rounded corners] ($(desc.north west)+(-0.2,1)$)  rectangle ($(comment.south east)+(-0.1,-0.1)$);

\node[inner sep=0pt]  (title_PAL) at (9,0) {\textbf{Positive Active Learning (PAL)}};

\node[inner sep=0pt,below=0.1cm of title_PAL] (desc_PAL)  {
\parbox{7.8cm}{\small\em Given inputs, choose if they are semantically related: \texttt{yes}/\texttt{no}}};

\node[inner sep=0pt, below left=0.6cm and 0.4cm of title_PAL] (q1title_PAL) {$\texttt{query}_{\texttt{1}}$};
\node[inner sep=0pt,below=0.1cm of q1title_PAL] (q1_PAL){\includegraphics[width=2cm,height=2cm]{figures/teaser/ILSVRC2012_val_00038928.jpg}};
\node[inner sep=0pt, below=0.1cm of q1_PAL] (and1) {$and$};
\node[inner sep=0pt,below=0.1cm of and1] (q1_PAL2){\includegraphics[width=2cm,height=2cm]{figures/teaser/ILSVRC2012_val_00008373.jpg}};
\node[inner sep=0pt,below=0.3cm of q1_PAL2] (option1_PAL)  {\parbox{2cm}{\centering$\texttt{response}_{\texttt{1}}$\\ yes}};

\node[inner sep=0pt,below right=0.6cm and -4cm of title_PAL] (q2title_PAL)  {$\texttt{query}_{\texttt{2}}$};
\node[inner sep=0pt,below=.1cm of q2title_PAL] (q2_PAL) {\includegraphics[width=2cm,height=2cm]{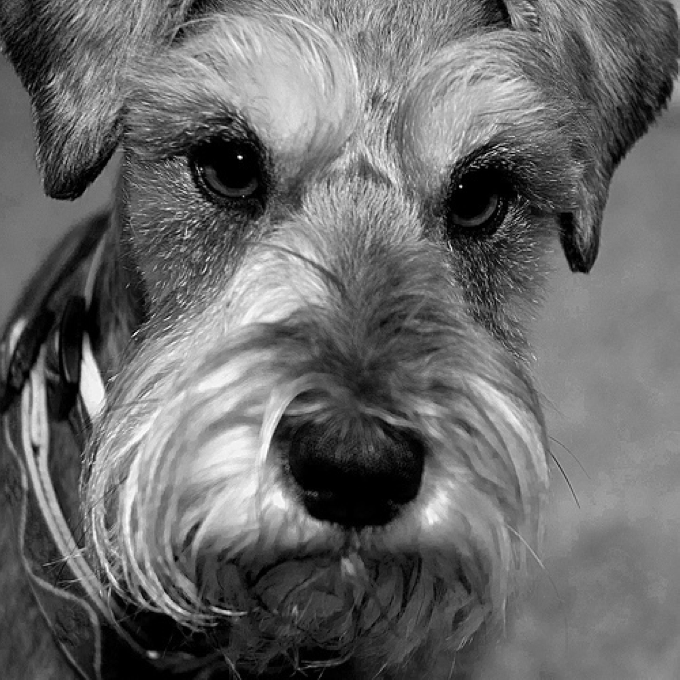}};
\node[inner sep=0pt, below=0.1cm of q2_PAL] (and2) {$and$};
\node[inner sep=0pt,below=0.1cm of and2] (q2_PAL2){\includegraphics[width=2cm,height=2cm]{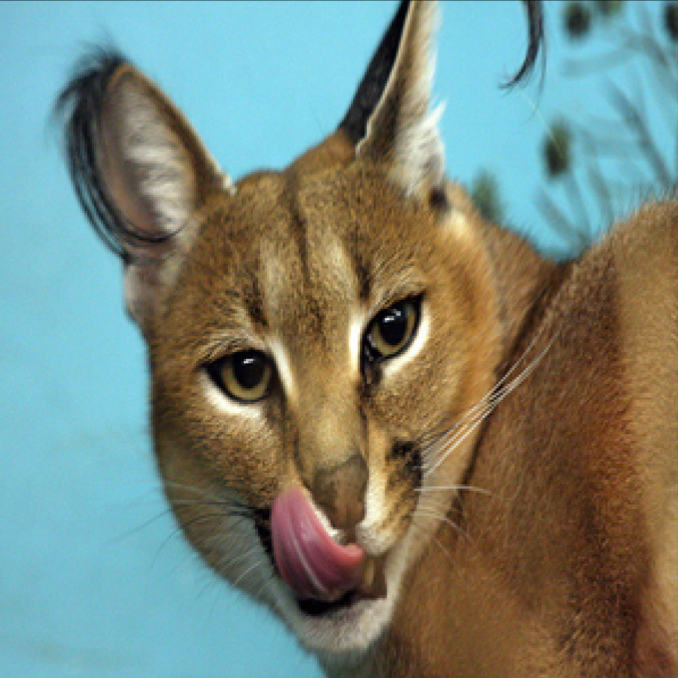}};

\node[inner sep=0pt,below=0.3cm of q2_PAL2] (option2_PAL)  {\parbox{2cm}{\centering$\texttt{response}_{\texttt{2}}$\\no}};

\node[inner sep=0pt,below right=-2.8cm and 1.4cm of and2] (recap){\includegraphics[width=4cm,height=5.75cm]{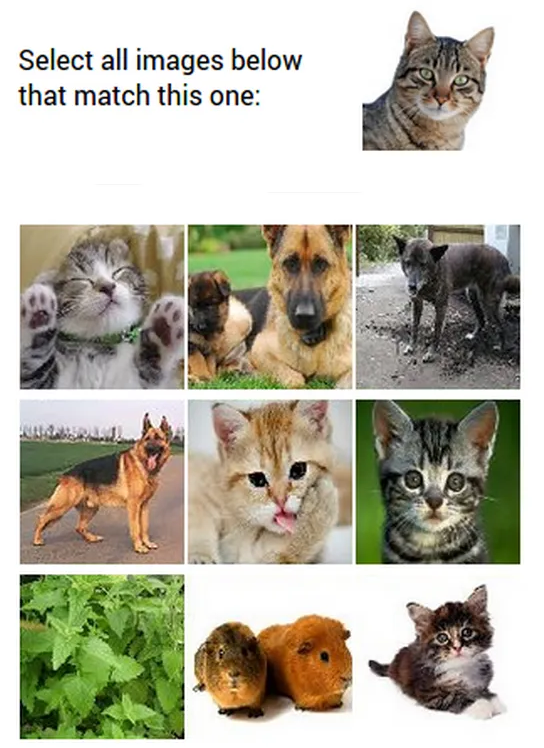}};

\node[inner sep=0pt,below=6.5cm of title_PAL,color=blue] (comment_PAL)  {\centering
Low-cost relationships information (reduced expertise)
};
\draw[black,thick,rounded corners] ($(desc_PAL.north west)+(-.8,0.6)$)  rectangle ($(comment_PAL.south east)+(1.6,-0.2)$);

\node[right=1cm of and2] (a) {\rotatebox{90}{\textit{ or (recaptcha)}}};
\node[below=1cm of a] (b) {};
\node[above=1cm of a] (c) {};

\draw (b) -- (a) -- (c);

\end{tikzpicture}
\end{adjustbox}
\vspace{.25em}
    \caption{Active Self-Supervised Learning  introduces PAL ({\bf right box}), an alternative to active learning ({\bf left box}) where the oracle is asked if a collection of inputs are semantically related or not. 
    As opposed to active learning, expert knowledge is reduced as one need not to know all the possible classes but only how to distinguish inputs from different classes. 
    PAL querying is flexible, as an illustrative example we exhibit an \`{a} la captcha version where a given input is presented along with a collection of other inputs, and the oracle can select among those inputs the positive ones.}
    \label{fig:pal}
\end{figure*}

\section{Introduction}

Learning representations of data that can be used to solve multiple tasks, out-of-the-box, and with minimal post-processing is one of the main goals of current AI research \cite{pan2010survey,lecun2015deep,foundation}. 
Such representations are generally found by processing given inputs through Deep neural Networks (DNs). 
The main question of interest around which contemporary research focuses on deals with the choice of the training setting that is employed to tune the DN's parameters. 
A few different strategies have emerged such as layerwise \cite{bengio2006greedy}, reconstruction based \cite{vincent2008extracting}, and more recently, based on Self-Supervised Learning (SSL) \cite{chen2020simple,misra2020self}. 
In fact, due to the cost of labeling and the size of datasets constantly growing, recent methods have tried to drift away from traditional supervised learning \cite{settles2011theories}. 
From existing training solutions, joint-embedding SSL has emerged as one of the most promising ones \cite{jepa}.
It consists in learning representations that are invariant along some known transformations while preventing dimensional collapse of the representation. 
Such invariance is enforced by applying some known Data-Augmentation (DA), e.g. translations for images, to the same input and making sure that their corresponding representations are the same.

Despite tremendous progress, several limitations remain in the way of a widespread deployment of SSL.
In particular, it is not clear how to incorporate {\em a priori} knowledge into SSL frameworks beyond the usual tweaking of the loss and DAs being employed, although some efforts are being made \cite{chen2020big,zheltonozhskii2022contrast,zhai2019s4l}.
Indeed, it is not surprising that vision-language pre-training has replaced SSL as the state-of-the-art to learn image representation \cite{gan2022visionlanguage}, as those models are better suited to incorporate information stemming from captions that often come alongside images collected on the Internet.


In this study, we propose to redefine existing SSL losses in terms of a similarity graph --where nodes represent data samples and edges reflect known inter-sample relationships.
Our first contribution stemming from this formulation provides a {\em generic framework to think about learning in terms of similarity graph}: it yields a spectrum on which SSL and supervised learning can be seen as two extremes. 
Within this realm, those two extremes are connected through the similarity matrix, and in fact can be made equivalent by varying the similarity graph.
Our second contribution naturally emerges from using such a similarity graph, unveiling an elegant framework to reduce the cost and expert requirement of active learning summarized by:
\begin{center}
\begin{minipage}{0.6\linewidth}
\centering
\em
Tell me who your friends are,\\ and I will tell you who you are.
\end{minipage}
\end{center}
Active learning, which aims to reduce supervised learning cost by only asking an oracle for sample labels when needed \cite{Settles2010,Dasgupta2011,Hanneke2014,Karzand2020}, can now be formulated in term of relative sample comparison, rather than absolute sample labeling.
This {\em much more efficient and low-cost approach is exactly the active learning strategy stemming from our framework}: rather than asking for labels, one rather asks if two (or more) inputs belong to the same classes or not, as depicted in \cref{fig:pal}. We coin such a strategy as {\em Positive Active Learning (PAL)}, and we will present some key analysis on the benefits of PAL over traditional active learning. We summarize our contributions below:
\begin{itemize}
    \item We provide a {\em unified learning framework} based on the concept of similarity graph, which encompasses both self-supervised learning, supervised learning, as well as semi-supervised learning and many variants.
    \item We derive a {\em generic PAL algorithm} based on an oracle to query the underlying similarity graph \cref{alg:pal}.
    The different learning frameworks (SSL, supervised, and so forth) are recovered by different oracles, who can be combined to benefit from each framework distinction.
    \item We show how PAL extends into an {\em active learning framework} based on similarity queries that provides low-cost efficient strategies to annotate a dataset (\cref{fig:pal}).
\end{itemize}
All statements of this study are proven in \cref{proof:section}, code to reproduce experiments is provided at \url{https://github.com/facebookresearch/pal}.

\begin{figure*}[t!]
    \centering
    \begin{minipage}{.45\linewidth}
    \includegraphics[width=\linewidth]{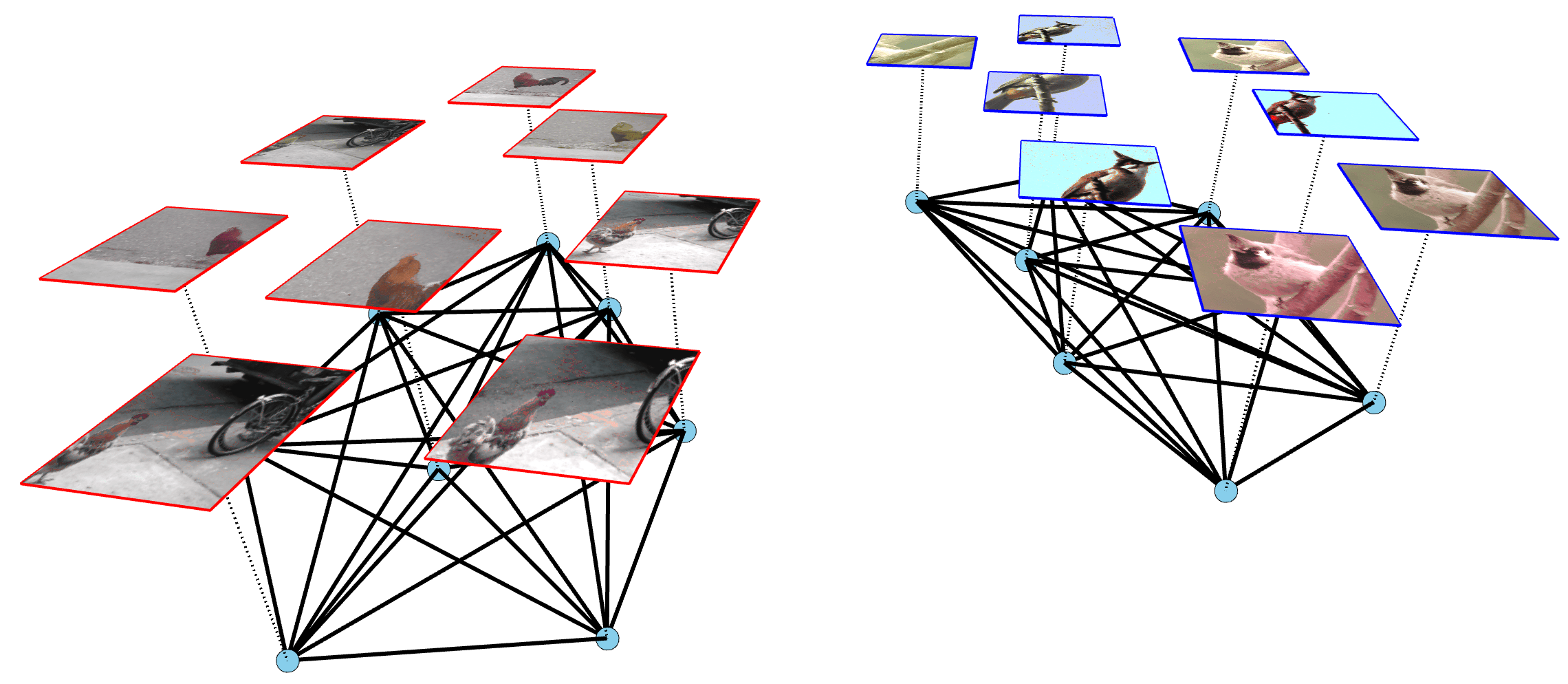}
    \end{minipage}
    \hspace{.05\linewidth}
    \begin{minipage}{.45\linewidth}
    \includegraphics[width=\linewidth]{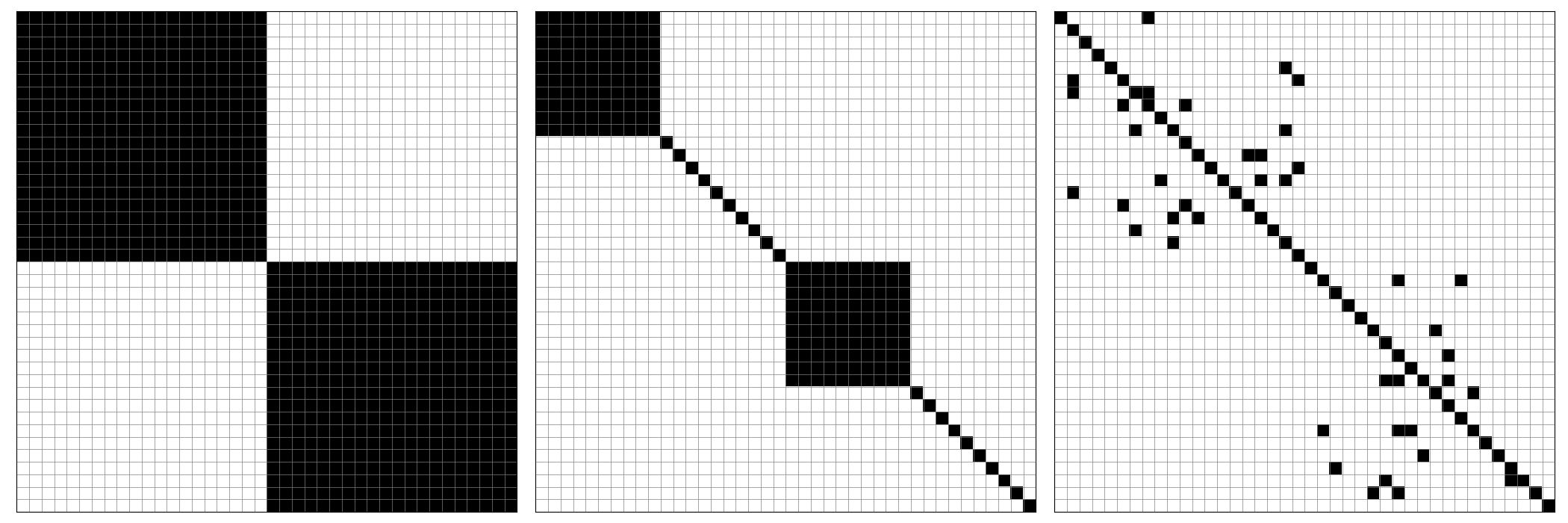}
    \hspace{.5\linewidth}
    \begin{minipage}{0.32\linewidth}\small
    \centering
    supervised
    \end{minipage}
    \begin{minipage}{0.32\linewidth}\small
    \centering
    semi-sup.
    \end{minipage}
    \begin{minipage}{0.32\linewidth}\small
    \centering
    PAL
    \end{minipage}
    \end{minipage}
    \vspace{.5em}
    \caption{\small {\bf Left:} Depiction of the ``knowledge graph'' arising from binary classification with supervised learning.
    Notice the two connected components, each corresponding to a single class. 
    Each sample is associated with a node of the graph (blue circle) and the known positive relation between samples is represented by an edge. This knowledge is summarized into the $\mG$ matrix depicted on the right.
    {\bf Right:} Examples of the $N \times N$ symmetric graph-adjacency matrices $\mG$ for the case of binary classification with supervised (same graph as on the left), semi-supervised and active learning. 
    Each nonzero entry $(\mG)_{i,j}$ represents the known positive relation between sample $i$ and $j$. 
    }
    \label{fig:G}
\end{figure*}

\section{Background on Self-Supervised Learning}
\label{sec:background}

This section provides a brief reminder of the main self-supervised learning (SSL) methods, their associated losses, and common notations for the remainder of the study.

A common strategy to learn a model in machine learning is to curate labeled examples $(\vx_n,y_n)_n$, and to learn a model that given $\vx_n \in \X \triangleq \R^D$ as input, outputs $y_n \in [C]$, hoping that this model will learn to recognize patterns and relations that generalizes to new, unseen input data. 
Yet, as the dataset grew larger, and annotating data has become a major bottleneck, machine learning has shifted its attention to learning methods that do not require knowledge of $y_n$.
SSL has emerged as a powerful solution to circumvent the need for expensive and time-consuming labeling.
It learns a embedding $f:\X\to\R^K$ for a small $K$ by enforcing either reconstruction properties, or some invariance and symmetry onto a learned representation. 
SSL also relies on a set of observations $\mX = \{\vx_n\}_{n=1}^{N} \in \R^{N\times D}$, yet instead of labels $y_n$, it requires known {\em pairwise positive relation} that indicates whether two samples are semantically similar or not. 
For simplicity, we shall focus on the joint-embedding framework, where those positive pairs are artificially generated on the fly by applying Data Augmentations (DA), e.g. adding white noise, masking, on the same input.
Let denote $\gT_1, \gT_2:\X\to\X$ the generators of two (random) DAs $\gT_1(\vx)$ and $\gT_2(\vx)$ from an input $\vx$, $f_{\theta}:\mathbb{R}^{D} \to \mathbb{R}^{K}$ the  parametric model to be learned, and
\begin{align}
\mZ^{(1)} \triangleq \begin{bmatrix}
f_{\theta}(\gT_1(\vx_1))\\
\vdots\\
f_{\theta}(\gT_1(\vx_N))\\
\end{bmatrix},
\mZ^{(2)} \triangleq \begin{bmatrix}
(f_{\theta}(\gT_2(\vx_1))\\
\vdots\\
(f_{\theta}(\gT_2(\vx_N))\\
\end{bmatrix},\label{eq:Z}
\end{align}
where $(\vz^{(1)}_n,\vz^{(2)}_n)$, the $n^{\rm th}$ row of $\mZ^{(1)}$ and $\mZ^{(2)}$ respectively, form the $n^{\rm th}$ positive pair associated to sample $\vx_n$. Using \eqref{eq:Z}, different SSL losses will employ different measures of invariance and dimensional collapse. 
Typically, the losses are minimized with gradient descent and backpropagation to learn $\theta$.

{\bf VICReg.}~With the above notations, the VICReg loss \cite{bardes2021vicreg} reads, with hyper-parameter $\alpha, \beta > 0$,
\begin{align}
\nonumber
&\gL_{\rm VIC} = \alpha \sum_{k=1}^{K}\relu\left(1-\sqrt{\mC_{k,k}}\right)+\beta \sum_{k\neq l}\mC^2_{k,l} 
\\&\quad+ \frac{1}{N}\|\mZ^{(1)}-\mZ^{(2)}\|_2^2,
\qquad\mC \triangleq \Cov(\begin{bmatrix} \mZ^{(1)}\\\mZ^{(2)} \end{bmatrix}).
\label{eq:VICReg}
\end{align}

{\bf SimCLR.}~The SimCLR loss \cite{chen2020simple} with temperature hyper-parameter $\tau > 0$ reads
\begin{align}
    \nonumber
    &\gL_{\rm Sim}=-\sum_{i=1}^{N}\frac{\mC_{i,i}}{\tau}+\log\left(\sum_{i\neq j}^{N}\exp\paren{\frac{\mC_{i,j}}{\tau}}\right),
    \\& \mC_{i,j} \triangleq \CosSim(\mZ^{(1)}, \mZ^{(2)})_{ij} \triangleq \frac{\scap{\vz^{(1)}_i}{\vz^{(2)}_j}}{\|\vz^{(1)}_i\| \,\|\vz^{(2)}_j\|},\label{eq:simCLR}
\end{align}

{\bf BarlowTwins.}~BarlowTwins \cite{zbontar2021barlow} is built on the cross-correlation matrix $\mC_{ij} = \CosSim(\mZ^{(1)}{}^\top, \mZ^{(2)}{}^\top)$, with the hyper-parameter $\lambda$ it reads
\begin{align}
    \label{eq:BT}
    \gL_{\rm BT} = \sum_{k=1}^K (1 - \mC_{ii})^2 + \lambda \sum_{i\neq j} \mC_{ij}^2.
\end{align}

{\bf Spectral Contrastive Loss.}~Finally, the spectral contrastive loss \cite{haochen2021provable} is theory-friendly proxy for SSL reading 
\begin{align}
    \label{eq:scl}
    \gL_{\rm VIC^2} = -2\scap{\mZ^{(1)}}{\mZ^{(2)}} + \frac{1}{N}\sum_{i\neq j} \scap{\vz^{(1)}_i}{\vz^{(2)}_j}^2.
\end{align}
In particular, as proven in \cref{proof:vic-2}, \eqref{eq:scl} recovers VICReg \eqref{eq:VICReg} when the ReLU-hinge loss is replaced by the mean-square error, hence the denomination VIC$^2$.

{\bf The Commonality between SSL Losses.}~%
All the above \cref{eq:VICReg,eq:simCLR,eq:scl,eq:BT} losses combine two terms: (i) a matching term between positive pairs, and (ii) a term to avoid collapse towards predicting a constant solution for all inputs.
(i) can take different forms such as the squared norm between $\mZ^{(1)}$ and $\mZ^{(2)}$ \eqref{eq:VICReg}, the opposite of their scalar product \eqref{eq:scl}, or of their cosine \eqref{eq:simCLR}, or the square norm between the centered-cosine and one \eqref{eq:BT}.
(ii) can also take different forms such as the infoNCE softmax \eqref{eq:scl}, or an energy that enforces richness of the learn feature, such as the variance-covariance regularization in \eqref{eq:VICReg} forcing the different coordinates of $f_\theta$ to be orthogonal \cite{cabannes2023ssl}.

While at face-value those losses seem distinct, they actually all simply consist and combine some variants of (i) and (ii), and even more importantly, they all rely on the exact same information of positive inter-sample relation for (i). This is exactly what the next \cref{sec:unify} will dive into, as a means to unify SSL losses, along with supervised learning methods.

\section{The Ubiquity of Similarity Graphs}\label{sec:unify}

The goal of this section is to unify SSL and supervised learning through the introduction of a special object: a {\em similarity graph}.

\subsection{The Graphs for (Self-)Supervised Learning}

Throughout this study, a similarity graph denotes a graph for which nodes represent data samples, and edges reflect similarity relationships. 
Formally, such a graph is expressed through its symmetric adjacency matrix $\mG \in \R^{N \times N}$, the semantic relation between inputs $i$ and $j$ being encoded in the real entry $\mG_{i,j}$. The remainder of this section will demonstrate how (i) SSL losses are implicitly based on a similarity graph (ii) how those losses tackle the supervised learning problem when provided a richer graph $\mG$.

{\bf Supervised learning.}~%
In addition to the $N$ input samples $\mX \in \R^{N\times D}$, supervised learning has access to paired labels $\vy \triangleq [y_1,\dots,y_N]$.
For clarity, we focus here on categorical labels  i.e. $y_n$ belongs to $\{1,\dots,C\}$ for $C$ the number of classes.\footnote{%
    While we focus here on classification for simplicity, our approach is easily extendable for generic problems involving a loss $\ell$ by defining the graph as $\mG_{ij} = -\ell(y_i, y_j)$. 
    In the classification, $\ell$ could be the zero-one loss $\ell(y_i, y_j) = \ind{y_i\neq y_j}$, and $\mG_{ij} \simeq 1 - \ell(y_i, y_j)$.
    See \cref{app:recovery} for details.
}
The one-hot encoding of $\vy$ will be denoted by the matrix $\mY \in \R^{N\times C}$.
In terms of the similarity graph $\mG$, the label-based relation becomes naturally encoded as $\mG_{i,j} = \ind{y_i\neq y_j}$, or equivalently
\begin{align}
    \label{eq:G_sup_easy}
    \mG(\mY) = \mY \mY^\top
\end{align}
A key observation that we must emphasize is that the graph $\mG$ almost explicitly encodes for the labels $\mY$, which will be made explicit with Theorem \ref{thm:recovery}.

{\bf Multiple Epochs with Data Augmentation.}~%
When DA is employed, and training is carried for $E$ epochs, the original $N$ input samples are transformed into $N\times E$ ``augmented'' samples. For more generality, since DA will also be used in SSL, let's denote by $A \in \mathbb{N}^*$ the number of augmentations --where here $A=E$. We now have the augmented dataset
\begin{align*}
    \mX^{(A)} &\triangleq [\underbrace{\gT(\vx_1),\dots,\gT(\vx_1)}_{\text{repeated $A$ times}} ,\dots, \gT(\vx_N),\dots,\gT(\vx_N)]^\top,
\end{align*}
where each $\gT$ has its own randomness. When available, i.e. for supervised learning, the corresponding ``augmented'' labels $\mY^{(A)}$ are given by repeating $A$ times each row of $\mY$, formally written with the Kronecker product $\mY^{(\sup)}\triangleq \mY \otimes \1_A$, and from that, we can now define the supervised dataset and associated graph extending \eqref{eq:G_sup_easy} to the case of multiple epochs and DA training
\begin{align}
    \label{eq:G_sup}
    \mX^{(\sup)}\triangleq \mX^{(E)},\qquad
    \mG^{(\sup)} \triangleq {\mY^{(\sup)}}^\top \mY^{(\sup)}.
\end{align}
The resulting graph \eqref{eq:G_sup} is depicted on the left part of \cref{fig:G}.

{\bf Self-Supervised Learning.}~%
SSL does not rely on labels, but on positive pairs/tuples/views generated at each epoch. 
Let us denote by $V$ the number of positive views generated, commonly $V=2$ for positive pairs as modeled in \eqref{eq:Z}.
With $E$ the total number of epochs, SSL produces $V \times E$ samples semantically related to each original sample $\vx_n$ through the course of training {\em i.e.} in SSL $A= V\times E$ while in supervised learning $A=E$. 
The total number of samples is thus $N\times E\times V$, defining the dataset and associated graph
\begin{align}
\mX^{(\ssl)} \triangleq\mX^{(V\times E)},\,
     \mG^{(\ssl)}_{i,j} = \Indic_{\{\floor{i/VE}=\floor{j/VE}\}},\label{eq:G_ssl}
\end{align}
where the associated similarity graph $\mG^{(\ssl)}$ --now of size $NEV\times NEV$-- captures if two samples were generated as DA of the same original input.

\subsection{Self-Supervised Learning on a Graph}

This section reformulates the different SSL losses through the sole usage of the similarity graph $\mG^{(\ssl)}$. 
To lighten notations, and without loss of generality, we {\em redefine} $\mX \in \R^{N\times D}$ to denote the full dataset, i.e. $N \leftarrow NEV$ with $\mX = \mX^{(\sup)}$ for supervised learning with $V \times E$ epochs, or with $\mX = \mX^{(\ssl)}$ in SSL with $E$ epochs with $V$ views for the SSL case.
The model embedding is shortened to $\mZ \triangleq f_\theta(\mX) \in \R^{N\times K}$ as per \cref{eq:Z}.

\begin{theorem}
\label{lemma:characterization}
VICReg \eqref{eq:VICReg}, SimCLR \eqref{eq:simCLR}, and BarlowTwins \eqref{eq:BT} losses can be expressed in term of the graph $\mG$ \eqref{eq:G_ssl}
\begin{align*}
    \\ \gL_{\rm VIC^2}(\mZ;\mG)=&\| \mZ\mZ^T  - \mG \|_F^2,
    \\ \gL_{\rm Sim}(\mZ;\mG)=&-\hspace{-0.2cm}\sum_{i,j\in[N]}\mG_{i,j}\log\paren{\frac{\exp(\tilde\vz_i^\top \tilde\vz_j)}{\sum_{k\in[N]} \exp(\tilde\vz_i^\top \tilde\vz_k)}},
    \\ \gL_{\rm BT}(\mZ;\mG)=& \norm{\tilde\mZ^\top \mG \tilde\mZ - I}^2,
\end{align*}
where $\mD = \diag(\mG \1)$ is the degree matrix of $\mG$; with $\tilde\vz \triangleq \vz / \norm{\vz}$ and $\tilde\mZ$ the column normalized $\mZ$ so that each column has unit norm.
\end{theorem}

In essence, SSL is about making sure that sample's representations match for samples that were deemed similar through the design of data augmentation.
As such, it is not surprising that one can rewrite SSL losses through the sole usage of the similarity graph.
From \cref{lemma:characterization}, the attentive observer would notice how VICReg is akin to Laplacian Eigenmaps or multidimensional scaling, SimCLR is akin to Cross-entropy and BarlowTwins is akin to Canonical Correlation Analysis; observations already discovered in the literature \cite{balestriero2022contrastive} and reinforced above. 

Beyond recovering such representation learning losses, our goal is to go one step further and to tie SSL and supervised learning through the lens of $\mG$, which follows in the next section.

\subsection{Self-Supervised is a G Away from Supervised}

What happens if one takes the different SSL losses, but replaces the usual SSL graph $\mG^{({\rm ssl})}$ with the supervised one $\mG^{({\rm sup})}$?

It turns out that the learned representations emerging from such losses are identical --up to some negligible symmetries that can be corrected for when learning a linear probe-- to the one hot-encoding of $\mY$.
To make our formal statement (\cref{thm:recovery}) clearer, we introduce the set of optimal representations that minimize a given loss:
\begin{align*}
        \gS_{\rm method}(\mG) \triangleq \argmin_{\mZ \in \R^{N \times K}} \gL_{\rm method}(\mZ;\mG),
    \end{align*}
where ``method'' refers to the different losses. 

\begin{theorem}[Interpolation optimum]
    \label{thm:recovery}
    When $K \geq C$, and $\mZ = f_\theta(\mX)$ is unconstrained (e.g. interpolation regime with a rich functions class), the SSL losses as per \cref{lemma:characterization} with the supervised graph \eqref{eq:G_sup} solve the supervised learning problem with:
    \begin{align*}
        \gS_{\rm VIC^2}(\mG^{(\sup)}) &= \brace{\mY \mR \midvert \mR \in \R^{C\times K}; \mR \mR^\top = \mI_C},\\
        \gS_{\rm Sim}(\mG^{(\sup)}) &= \brace{\mD\mY\mR\mM^{-1}\midvert \mD \in \diag_+, \mR \in O},\\
        \gS_{\rm BT}(\mG^{(\sup)}) &= \brace{\mY\mR \mD \midvert \mD \in \diag_+, \mR \in O},
    \end{align*}
    where $\mR \in O$ means that $\mR$ is a rotation matrix as defined for the VICReg loss, $\diag_+ = \diag(\R^N_+)$ are the set of diagonal matrix with positive entries, i.e. renormalization matrices, and $\mM$ is a matrix that maps a deformation of the simplex into the canonical basis.
    Moreover, provided class templates, i.e. $C$ data points associated with each of the $C$ classes, $\mY$ is easily retrieved from any methods and $\mZ \in \gS_{\rm method}$.
\end{theorem}

In essence, \cref{thm:recovery} states that SSL losses solve the supervised learning problem when the employed graph $\mG$ is $\mG^{(\sup)}$.
Moreover, the matrix $\mD$ appearing in \cref{thm:recovery} captures the fact that SimCLR solutions are invariant to rescaling logit and is akin to the cross-entropy loss, while BarlowTwin is invariant to column renormalization of $\mZ$ and is akin to discriminant analysis.
Lastly, VICReg might be thought of as a proxy for the least-squares loss.
At a high-level, \cref{thm:recovery} suggests fruitful links between spectral embedding techniques captured in \cref{lemma:characterization} and supervised learning.
We let for future work the investigation of this link and translation of spectral embedding results in the realm of supervised learning.

While \cref{thm:recovery} describes what we have coined as the ``interpolation optimum'', i.e. solution in the interpolation regime with rich models, we ought to highlight that classical statistical learning literature analyzes losses under the light of ``Bayes optimum'', i.e. solutions in noisy context-free setting \cite{Bartlett2006}.
Those Bayes optima do not make as much sense for losses that intrinsically relate different inputs, yet for completeness we provide such a proposition on Bayes optimum in \cref{app:bayes}.

\section{PAL: Positive Active Learning}
\label{sec:PAL}

Now that we demonstrated how one should focus on the graph $\mG$, rather than the (self-)supervised loss, we turn our focus into getting that graph $\mG$. 
In particular, we propose an active learning framework that discovers $\mG$ through efficient, low-cost queries.

\subsection{One Framework to Rule Them All}

From our understanding (\cref{thm:recovery}), the difficulties of both supervised learning and SSL are the same: they need a correct graph $\mG$, i.e they need to identify samples that are semantically similar, either through label annotations or through the right design of DA.

\begin{algorithm}[ht]
\KwData{$\mX\in\R^{N\times D}$; unknown graph $\mG = \mG^{(\sup)}$.}
\KwResult{Embedding $f_\theta:\R^D \to \R^K$.}
Initialization: weights $\theta_0$, scheduler $(\gamma_t)$; $T\in\N$;\\
\For{$t \in [T]$}{
Collect $I_t, J_t\leftarrow$ from {\color{blue} sampler};\\
Collect $(\mG_{ij} = \ind{y_i=y_j})_{(i,j)\in I_t}$ from {\color{blue} labelers};\\
Update $\theta_{t+1} \leftarrow \theta_t - \gamma_t \nabla_{\theta} \mL(\theta_t; \mG, I_t, J_t)$.
}
\caption{PAL framework with {\color{blue} oracle}}
\label{alg:pal}
\end{algorithm}

Our framework suggests a generic way to proceed, having fixed the samples $\mX$ in advance, and without much {\em a priori} knowledge on the similarity graph $\mG$.
In an active learning spirit, one would like to design a query strategy to discover $\mG$, and an update rule for the learned parameter $\theta$.
To ground the discussion, let us focus on VICReg.
The variance-covariance term can be rewritten with $R(a, b) = (a^\top b)^2 - \norm{a}^2 - \norm{b}^2$, this leads to the formula, proven in \cref{proof:vic-2},
\begin{gather}
    \gL_{\rm VIC^2}(\theta; \mG, I, J) = \sum_{(i,j)\in I} \mG_{i,j} \norm{f_\theta(\vx_i) - f_\theta(\vx_j)}^2
    \\+ \sum_{(i,j)\in J} R(f_\theta(\vx_i), f_\theta(\vx_j)),
    \label{eq:vic_sgd}
\end{gather}
where $I = J = [N]^2$.
An oracle would typically consider two small sets of indices $I, J \subset [N]^2$, asks labelers to provide $\mG_{ij}$ for $i,j\in I$, and, given a step size $\gamma_t$, update the weights with
\begin{align}
    \theta_{t+1} = \theta_t - \gamma_t \nabla_{\theta} \mL(\theta_t; \mG, I, J),
    \label{eq:weight_update}
\end{align}
which could be performed with the sole access to $(\mG_{ij})_{(i,j)\in I}$.
The pairs in $J$ are used to avoid dimensional collapse, and in particular for the VICReg loss, to compute the variance-covariance regularization term.
The complete picture leads to PAL, \cref{alg:pal}.
A particularly useful features of SGD for active learning is its robustness to labeling noise \cite{Cabannes2022active}.
In other terms, {\em \cref{alg:pal} is robust to noise in the query answers}.

We will now dive more in-depth into two variants of oracles: passive and active ones. As we will see, passive oracles can recover traditional SSL as special cases, but will be much less efficient in learning good representation than active strategies.

\subsection{Passive Oracles}

Passive variations of the PAL algorithm consist in fixing the oracle behavior at iterations $t\in [T]$ before starting the training. This formulation, under which the oracle does not leverage any information collected along training, recovers both SSL and supervised learning, based on the different querying strategies.

{\bf Self-Supervised Oracle.}~%
Probably the simplest oracle to describe is the one corresponding to the SSL strategy.
The original VICReg algorithm \cite{bardes2021vicreg} is made of $t$ gradient updates over $T = N_0 E$ iterations with $N = N_0 V E$ samples, where $N_0$ is the number of original samples, $E$ is the number of epochs, $V$ the number of views.
At time $t\in[T]$, $I_t$ is chosen as $\brace{(2t+1, 2(t+1))}$, describing a positive pairs generated on the fly from one original sample $\vx_i$ for $i = t \,\operatorname{mod. }\, N_0$; and $J_t$ is chosen as some mini-batch to estimate the covariance matrix of the features at the current epoch.
Because it has been built to remove human feedback, SSL actually does not need to ask for labelers to query $\mG_{s,s+1}$ (where $s = 2t+1$), since it is known by construction that those entries are going to be one.

{\bf Supervised Oracle.}~When it comes to a supervised learning oracle, the supervised learning loss provided in \cref{lemma:characterization} --which is known to recover $\mY$ (given class templates) as per \cref{thm:recovery}-- is easily minimized with gradient descent based on \eqref{eq:vic_sgd}.
Hence a simple oracle to solve the supervised learning problem based on stochastic gradient descent: at time $t$, consider a random pair of indices $(i_t, j_t)$ and set $I_t = J_t \leftarrow \brace{(i_t,j_t)}$.
The querying of $\mG_{i_t,j_t}$ can either be done on the fly, or if the dataset is already annotated, it can be deduced from the knowledge of $\mG_{i_t, j_t} = \ind{y_{i_t} = y_{j_t}}$.

\begin{algorithm}[ht]
    SSL oracle: \\
    \hspace{.3cm} Sampler: $I_t = \brace{(i_{2t+1}, i_{2(t+1)})}$, $J_t$ a minibatch,\\
    \hspace{.3cm} Labeler: $\mG_{2t+1,2(t+1)} = 1$.\\
    Supervised oracle: \\
    \hspace{.3cm} Sampler: $I_t = J_t = \brace{(i_t, j_t)}$ random in $[N]^2$,\\
    \hspace{.3cm} Labeler: $\mG_{i,j} = \ind{y_i = y_j}.$
    \caption{Passive Oracle Specifications}
    \label{alg:passive}
\end{algorithm}

{\bf Theoretical Remarks.}~%
Remarking that \eqref{eq:vic_sgd} is an unbiased formulation of VICReg, in the sense that
\[
    \gL_{\rm VIC^2}(\mZ) = \E_{I,J\sim \uniform{[N]^2}}\bracket{\gL_{\rm VIC^2}(\mZ; I, J)}.
\]
As a consequence, when $\theta \mapsto \norm{f_\theta(\mX) f_\theta(\mX)^\top - \mG}^2$ is strongly convex, \cref{alg:active} with either the self-supervised or the supervised oracle will converge to the minimizer of the VICReg loss in $O(1/T)$ \cite{Bubeck2015}.
Moreover, while this results holds for the empirical loss with resampling, it is equally possible to get a similar result for the minimization of the infinite-data (aka population) version of the VICReg loss and the recovery of the ideal embedding representation, when performing a single pass over the data.
In particular, by making sure that $J$ only charges pairs $(i, j)$ for $i$ and $j$ in two disjoint  subsets of $[N]$, one can prove convergence rates in $O(1/N)$ (Theorem 3 in \cite{cabannes2023ssl}).

Moreover, because the VICReg loss in \cref{thm:recovery} is nothing but a matrix factorization problem, one can directly translate results from this literature body into PAL. In particular, recent works have derived theoretical results regarding the matrix factorization problem based on toy models of neural networks, which might be plugged directly in here to claim theoretical results about the soundness of the PAL algorithm with neural networks \cite{Ye2021,Du2018,Jian2022}.
Since those results hold for any graph $\mG$, such results directly apply to both SSL and supervised learning, highlighting how PAL jointly derives results for SSL and supervised learning. 

\begin{figure}[t]
    \centering
\begin{tikzpicture}
\node[inner sep=0pt]  (title) at (0,-.1) {Level lines of $\ve_3^\top f_\theta$ at snapshots (learned with VICReg)};

\node (fig11) at (-3, -1.25)  {
    \includegraphics[width=.2\linewidth]{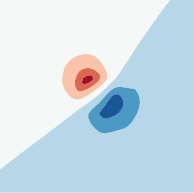}
};

\node[right=0.2cm of fig11] (fig12)  {
    \includegraphics[width=.2\linewidth]{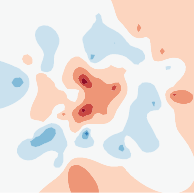}
};

\node[right=0.2cm of fig12] (fig13)  {
    \includegraphics[width=.2\linewidth]{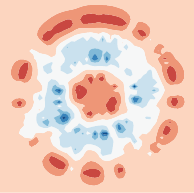}
};

\node[right=0.2cm of fig13] (fig14)  {
    \includegraphics[width=.2\linewidth]{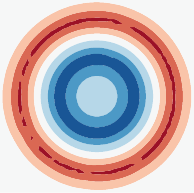}
};

\node[inner sep=0pt]  (title2) at (0,-2.5) {Optimal linear probe $w^\top f_\theta$ for downstream task};

\node (fig1) at (-3, -3.65)  {
    \includegraphics[width=.2\linewidth]{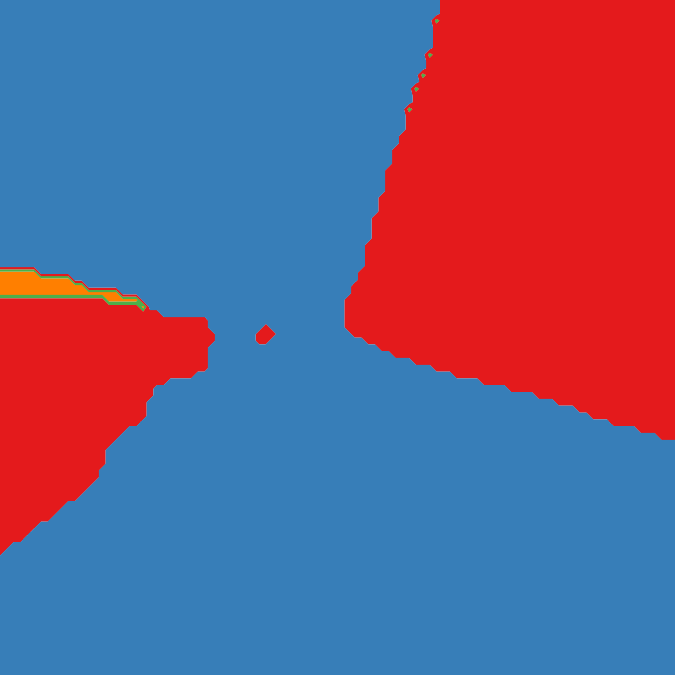}
};

\node[right=0.2cm of fig1] (fig2)  {
    \includegraphics[width=.2\linewidth]{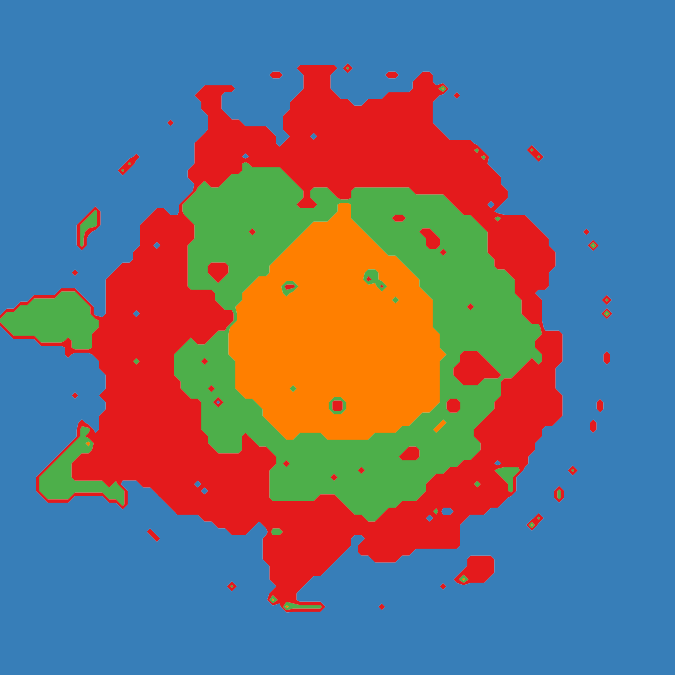}
};

\node[right=0.2cm of fig2] (fig3)  {
    \includegraphics[width=.2\linewidth]{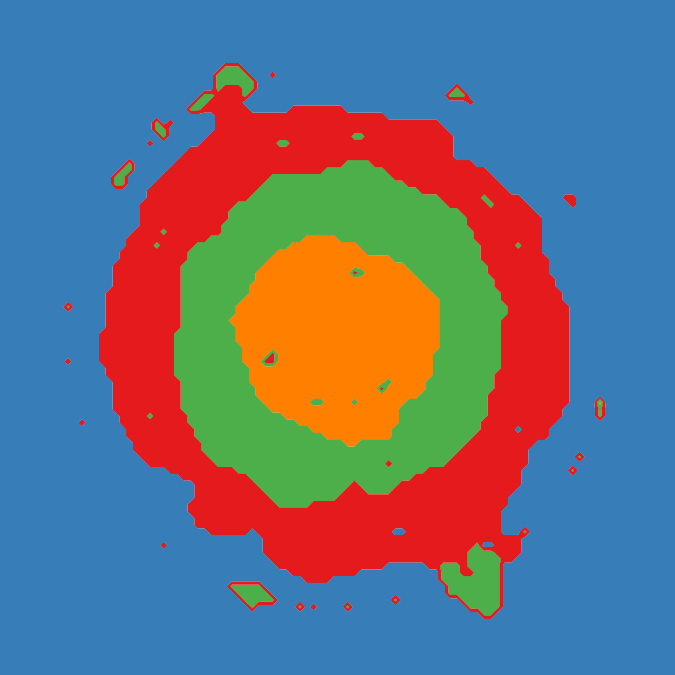}
};

\node[right=0.2cm of fig3] (fig4)  {
    \includegraphics[width=.2\linewidth]{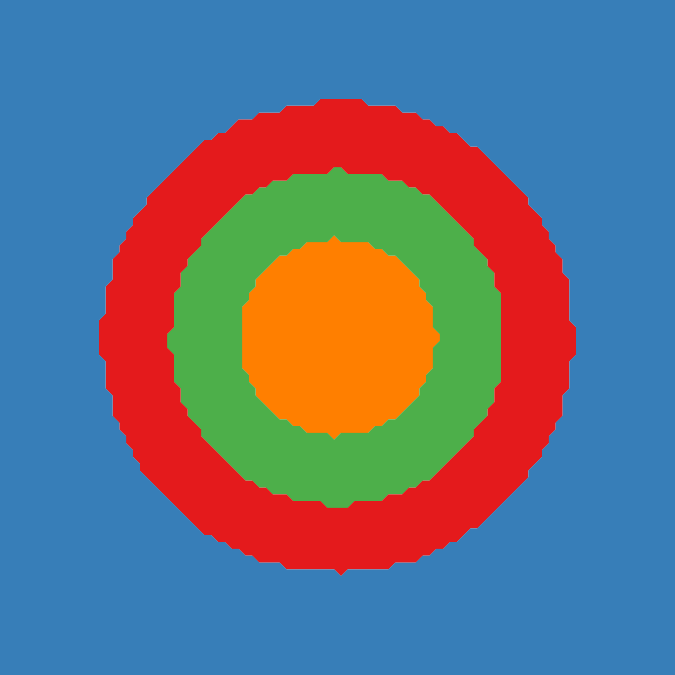}
};

\node (main) at (0, -7) {\includegraphics{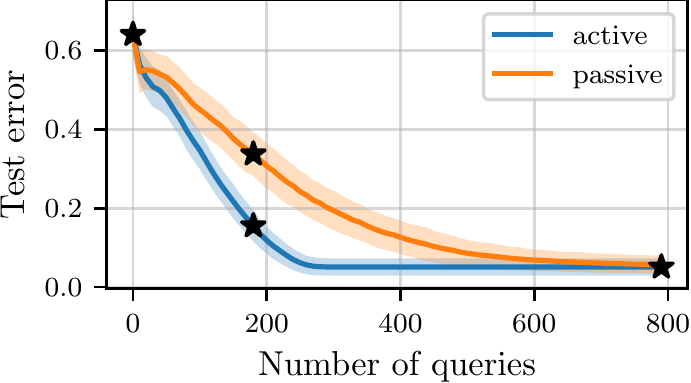}};

\draw[<-, very thick, draw opacity=.5] (fig1.south) -- ([shift=({-2.15,3.65})]main.south); 
\draw[<-, very thick, draw opacity=.5] (fig2.south) -- ([shift=({-.95,2.35})]main.south); 
\draw[<-, very thick, draw opacity=.5] (fig3.south) -- ([shift=({-.95,1.7})]main.south); 
\draw[<-, very thick, draw opacity=.5] (fig4.south) -- ([shift=({3.22,1.25})]main.south); 

\draw[->, draw opacity=.25] ([shift=({0,.3})]fig11.south) -- ([shift=({0,-.1})]fig1.north); 
\draw[->, draw opacity=.25] ([shift=({0,.3})]fig12.south) -- ([shift=({0,-.1})]fig2.north); 
\draw[->, draw opacity=.25] ([shift=({0,.3})]fig13.south) -- ([shift=({0,-.1})]fig3.north); 
\draw[->, draw opacity=.25] ([shift=({0,.3})]fig14.south) -- ([shift=({0,-.1})]fig4.north); 

\end{tikzpicture}
    \caption{Comparison the active oracle of \cref{alg:active} and the passive supervised one of \cref{alg:passive}.
    Given $q$ queries made, and the consequent reconstructed graph $\mG_q$, we learn $f_{\theta_t}:\X\to\R^C$ by minimizing $\gL_{\rm VIC^2}$, and plot the downstream mean-square error of the optimal a linear classifier $w^\top f_{\theta_t}$ for the best $w\in \R^C$.
    Here $\X = \R^2$, and $y\in[4]$ spans four concentric circles (represented by the blue, red, green and orange classes), $N=100$, query batches are chosen of size 10 and results are average over 100 trials (standard deviations being represented by the colorized regions). 
    Snapshots at different points on the curve show the third coordinates of the reconstructed $f_{\theta_t}$, and the ideal linear classifier that can be learned based on this embedding.
    }
    \label{fig:active}
\end{figure}

\subsection{Active Oracles}
Seen through the eyes of PAL, supervised and SSL --which employ passive querying-- can be improved by refining the oracle to choose the next indices $I_t$ and $J_t$ to process at time $t$.

{\bf Low-Cost and Efficient Active Learning.}~%
A crucial point of this study is that the active learning framework stemming from PAL differs fundamentally from classic active learning.
In the latter, at time $t$, one asks for a fresh label $y_{i_t}$ for some chosen index $i_t$.
Instead, PAL considers a batch of data $I_t$ and asks for pairwise comparisons $\ind{y_i\sim y_j}$ for $(i, j) \in I$.
Rather than asking labelers for fine-grained labels, such as ``caracal'' or ``numbfish'' on ImageNet, PAL would rather asks labelers if two images are related, or even to spot outliers in a set of images compared to a template, as illustrated on \cref{fig:pal}.\footnote{This ``spot-the-outliers'' strategy is formalized with $I_t = \{(i_t, j)\,\vert\, j \in \tilde{I}_t\}$ for $i_t$ representing the class template, and $\tilde{I}_t$ capturing the batch of data to spot outliers in.}
This is particularly useful when the cost of spotting a few outliers in a batch of $M$ images is much less costly than annotating $M$ data points.
On such instances, Criteo engineers found that batches of 15 images was a sweet spot in terms of labeling efficiency \cite{criteo}; while ImageNet was annotated by querying images on search engines, and spotting outliers among the results \cite{ImageNet}.
Meanwhile, reCaptcha (illustrated on \cref{fig:pal}) is said to have helped annotate millions of images \cite{captcha}.
We refer the curious reader to \cite{Simard2017} and references therein regarding the design of efficient user interfaces for those labeling tasks.

{\bf Zoo of Active Learning Strategies.}~%
By introducing PAL, we open a way to match the practice of active learning and its theory through a grounded framework that encompasses current heuristics to annotate big chunks of data.
While the optimal oracle depends on the problem idiosyncrasies, as well as the labeling cost model, the vast literature body on active learning provides many heuristics to design sophisticated or intricate oracles under different structural assumptions.
One could query information based on the least certain predictions \cite{Hanneke2014,Ash2020}; based on the distance to the decision boundaries \cite{pruning2022}; by comparing predictions made by different methods in an ensemble \cite{Bilgic2010active,ensemble2021}; or by finding the queries whose answers are the most likely to modify the current guess for $f_\theta$ \cite{Yang2016active,Knuth1977,Karzand2020}.
We refer the curious reader to Appendix \ref{app:active} for further discussion on the matter.
Throughout reviews, adaptations to PAL, ablation studies and comparisons on different benchmarks of those strategies is left for future work.

{\bf PAL {\em \`a la} Captcha.}~%
A natural and easy property to leverage in order to build active learning strategies is the fact that the $N^2$-entry matrix $\mG$ is actually derived from the $NC$-entry matrix $\mY$.
In particular, one can recover the full graph $\mG = \mG^{(\sup)}$ with less than $NC$ pairwise queries, and in the best case only $N$ queries --compare this to the $N^2$-entries that are queried by the supervised learning oracle.
This idea is captured formally with the oracle described in \cref{alg:active}, where the matrix $\mQ$ remembered past queries, and illustrated on \cref{fig:active}.
At time $t$, this oracle chooses to query against the class with the least numbers of known instances, and choose $M$ data points, ask if they match this class, and update known labels as a consequence.
An advantage of the query strategy of \cref{alg:active} is that one can stop at any time $t$ and have a balanced labeled dataset to learn with.

\begin{algorithm}[ht]
    \KwData{Class templates $(\vmu_1, \cdots, \vmu_C) \in \X^C$, $\mQ \in \R_{N\times C}$ initialized at zero.}
    Choose the class with least known examples $j = \argmin_j {\bf 1}^\top \mQ_t \ve_j \in [C]$;\\
    Collect pairwise comparison $\mQ_{ij} \leftarrow \ind{\vx_i \sim \mu_j}$ for $i$ in a batch $B \subset [N] \setminus \mK_t$ where $\mK_t$ remove queries with known results based on $\mQ_t$;\\
    Sampler: $I_t = J_t$ all the new entries deduced in $\mG$.\\
    Labeler: Human feedback $\mQ_{ij}$; deduction to fill $\mG$.
    \caption{Oracle {\em \`a la} Captcha}
    \label{alg:active}
\end{algorithm}

The basic \cref{alg:active} can be improved in several ways.
First, class templates can be deduced based on initial queries: the first data point $\mu_1 = \vx_1$ provides a first class template; after querying $\ind{\vx_2\sim \vx_1}$ if the answer is negative, $\mu_2 = \vx_2$ provides a second class template (otherwise it is part of class one); so forth and so on (if $\ind{\vx = \mu_1} = \cdots = \ind{\vx=\mu_k} = 0$, set $\mu_{k+1} = \vx$).
Those templates could be refined during training by defining the templates as the example the most at the center of the classes examples with some well-thought notion of distance (either in the input space or the embedding space).
Second, when classes are unbalanced and class probabilities are roughly known, one should rather choose $y(t)$ to be the class that minimizes the number of known examples in this class divided by the probability of this class.
Third, if $C$ the number of classes is small, random sampling of the batch $B$ will work well enough.
Yet, when $C$ is big, random sampling will mainly lead to negative observations and too few positive ones.
In this situation, the algorithm is improved by training a classifier based on known labels at time $t$ (eventually incorporating unlabeled data with semi-supervised learning techniques), and querying labels that were classified as the same class.
Finally, to avoid only getting negative pairs on datasets with many classes, one could leverage hierarchy in the labels: if dealing with the classes of ImageNet, one can first ask reviewers coarse-grained information, e.g. flag pictures that are not fishes; before going deeper in the taxonomy.

\section{Experiments}
\label{sec:experiments}

This section provides experimental details to validate the various theoretical results derived in previous sections. In order to remove confounding effects linked with architecture, optimization, data curation and other design choices that might impact the different empirical validation we focus here on closed-form solution based on kernel methods with synthetic dataset.
Further real-world empirical validations are provided in \cref{app:expe}.
In particular, \cref{tab:nnclr} reminds the reader how NNCLR \cite{dwibedi2021little} succeed to beat state-of-the-art SSL methods on ImageNet thanks to an active labeler oracle, which defines positive pairs as nearest neighbors in the embedding space $f_{\theta_t}(\X)$.

\begin{table}[t]
    \centering
    \begin{tabular}{|c|c|c|c|}
    \hline
         Modality & Oracle & \@1 accuracy & \@5 accuracy  \\
    \hline
         Passive & SimCLR \cite{simclrv2} & 71.7 & - \\
         Passive & VICReg \cite{bardes2021vicreg} & 73.2 & 91.1 \\
         {\bf Active} & NNCLR \cite{dwibedi2021little} & {\bf 75.6} & {\bf 92.4}\\ 
    \hline
    \end{tabular}
    \vspace{1em}
    \caption{Best known performance on ImageNet for state-of-the-art SSL methods.
    Notice how NNCLR \cite{dwibedi2021little} derives states of the art performance on ImageNet thanks to an active rule for labelers in \cref{alg:pal}, which consists in defining positive pairs as nearest neighbors in the embedding space as detailed in \cref{alg:nnclr}.
    This rule allows to beat the passive strategy that are provided by SimCLR and VICReg.
    }
    \label{tab:nnclr}
\end{table}

\begin{figure}[ht]
    \centering
    \includegraphics{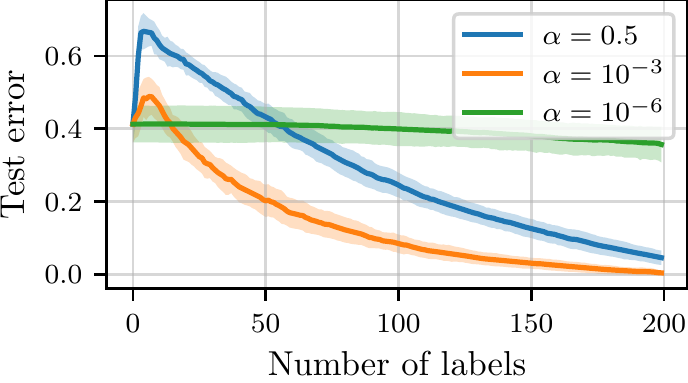}
    \caption{
    A major motivation of this paper is to be able to add prior information on sample relationships in SSL methods, and more in particular, to have a simple way to leverage known labels.
    We do by considering $\mY$ containing one-hot encoding of known labels, and rows being zero otherwise and the mixed graph
    \(
        \mG = (1-\alpha) \cdot \mG^{(\ssl)} + \alpha \cdot \hat\mY\hat\mY^\top.
    \)
    The setting is the same as \cref{fig:setup_toy} with $N=200$ and two augmentations per sample.
    When zero labels are known (left of the plot), we are in the full SSL regime, while when all the 200 labels are known (right of the plot), we recover supervised learning performance.
    When few labels are given the effect of the supervised graph can be counterproductive if the mixing coefficient $\alpha$ is too big.
    However, when mixed properly, adding prior label information in SSL methods allows to improve performance.
    }
    \label{fig:interpolation}
\end{figure}

Kernel methods are rich ``linear'' parametric models defined as $f_\theta = \theta^\top \phi(\vx)$, for $\phi(\vx)$ and $\theta$ belonging to a separable Hilbert space ${\cal H}$.
Because those model can approximate any function \cite{Micchelli2006}, it is important to regularize $\theta$ in practice, either with early stopping in SGD, or with Tikhonov regularization, which can be written as $\lambda\trace(\mZ^\top \mK^{-1}\mZ)$ where $\lambda > 0$ is a regularization parameter and $\mK \in \R^{N\times N}$ is the kernel matrix defined as $\mK_{ij} = k(\vx_i, \vx_j) = \phi(\vx_i)^\top\phi(\vx_j)$.
In this setting, rather than matching the top of the spectral decomposition of $\mG$, the solution recovered by VICReg amounts to the top spectral decomposition of $\mG - \lambda \mK^{-1}$ \cite{cabannes2023ssl}.
This allows to compute the ideal representation of $f_\theta$ in closed-form given any graph $\mG$ based on the regularized kernel model $f_\theta = \theta^\top \phi(\vx)$, hence ablating the effects that are unrelated to the theory described in this study.
In this controlled setting, the superiority of active algorithms is undeniable, and illustrated on \cref{fig:active}, where we illustrate the optimal downstream error one can achieve with linear probing of the minimizer $f_\theta$ of the VICReg loss.
Experimental details and more extensive validations are provided in \cref{app:expe}: in particular, the use of non-contrastive versus contrastive graphs, i.e. that set $\mG_{ij} = -1$ when $y_i \neq y_j$, is studied on \cref{fig:contrastive}; the ability to incorporate label knowledge in SSL methods is the object of \cref{fig:interpolation}; robustness to noise is shown on \cref{fig:noise}; and relations between test error and the number of connected components of the reconstructed $\mG$ is analyzed on \cref{fig:comp}.

\section{Conclusions}

This work introduces PAL, a learning framework that revolves around the central concept of similarity graph.
We first showed how similarity graphs are the implicit backbone of self-supervised learning methods, and how this concept extends to tackle supervised learning problems.
This observation does not solely unveil a rich learning framework, but also provides a single algorithm based on a querying oracle that can describe both SSL and supervised learning techniques, opening the way to new oracles that benefit from techniques stemming from both the supervised and self-supervised learning literature.
Finally, PAL leads to an efficient formalization of active learning as performed in practice to annotate large datasets, potentially enabling fruitful exchanges between the practice and the theory of active learning.
Promising directions for future works include empirical validations on large-scale datasets, as well as theoretical study of the newly introduced active learning framework.

{\small
\bibliographystyle{style/ieee_fullname}
\bibliography{references}
}

\onecolumn

\appendix
\section{Active Learning Algorithms}
\label{app:active}

{\bf NNCLR oracle.}~%
State-of-the-art the year of its release, \cite{dwibedi2021little} proposes a variation of SSL, that does not need human feedback but does update the similarity graph based on past observation.
This furnishes strong evidence for the usefulness of active learning algorithms, even without human feedback.
The NNCLR oracle consists in setting $I_t = J_t$ equals to some minibatch at time $t$, but to labels positive pairs in $\mG_{ij}$ only for nearest neighbors, i.e.
\[
    \mG^{(t)}_{ij} = \ind{\vz_i\in{\cal N}(\vz_j, I)} \triangleq \ind{ \norm{\vz_i - \vz_j} = \min_{j\in I_t} \norm{\vz_i - \vz_k}}.
\]
We describe the corresponding active oracle in \cref{alg:nnclr}.

\begin{algorithm}[ht]
    NNCLR oracle: \\
    \hspace{.3cm} Sampler: $I_t = J_t$ is some minibatch,\\
    \hspace{.3cm} Labeler: $\mG_{i,j} = \ind{\vz_i\in{\cal N}(\vz_j; I)}$ where ${\cal N}(\vz; I)$ design the nearest neighbor of $\vz$ in the batch $I$.
    \caption{Active Oracle with No Human Feedback as per \cite{dwibedi2021little}}
    \label{alg:nnclr}
\end{algorithm}

{\bf The New Rise of Active Learning.}~%
Recently, active learning has become a focus of the machine learning community when training big models, as those models performance are known to depend on the order they process data \cite{curriculum}, as well as in the percentage of data of different type they ingests \cite{llama}.
In particular, the best paper award at NeurIPS last year \cite{pruning2022} has suggested that the distance to the decision boundary could be leveraged smartly to reduce the number of data needed by current AI algorithms to train state of the art models (as compared to the scaling law of \cite{scaling2020}).
We describe their sampling oracle in \cref{alg:boundary}.
Note that in the original paper, they query the exact labels of the selected points, while PAL only needs to query pairwise comparison.
In the meantime, \cite{ensemble2021} suggested using ensemble active learning, while \cite{Ash2020} suggested a model based uncertain predictions through gradient computation in deep learning models.

\begin{algorithm}[ht] 
    Perform k-means clustering on $\mZ_t = f_{\theta_t}$.\\
    For each unlabeled points, compute cosine distance to its cluster center.\\
    \eIf{Few examples have been labels}{$I_t=J_t\leftarrow$ points near cluster centers,}{$I_t=J_t\leftarrow$ points far from cluster centers.}
    \caption{Active Oracle with Data Pruning as per \cite{pruning2022}}
    \label{alg:boundary}
\end{algorithm}

{\bf From Coarse to Fine-grained Query.}
While active learning usually assumes that the cost of answering any questions is constant, in practice, some queries might be easier to answer than others.
For example, if a child has never seen some objects, such as a sophisticated designer chair, they might not easily provide pairwise comparison regarding those objects, e.g. they would be puzzled by the designer chair, and would hesitate to say that this is a chair.
Similarly, it might be easier for human labelers to recognize attributes in an image, e.g. sandy fur, desert background, tufted ear, feline; rather than precise species, such as ``caracal''.
This has been the basis for weakly supervised learning \cite{Cour2011,Ratner2020}.
It has also motivated some bandit models, such as \cite{Cesa-Bianchi2006,Fiez2019}.
More generally, it suggests that one could efficiently learn by first querying weak, coarse-grained information, before refining queries to get precise, fine-grained feedback.
We illustrate this high-level idea with \cref{alg:hierarchical}.

\begin{algorithm}[ht]
    Samplers: Your favorite active learning sampler.\\
    \eIf{$f_{\theta_t}$ has not formed strong opinions on clusters}{Labelers: Human feedback for coarse-grained information (e.g. click on all animals with fur...),}{Labelers: Human feedback with fine-grained information (e.g. click on fishes that match a precise species).}
    \caption{Active Oracle with Hierarchical Taxonomy}
    \label{alg:hierarchical}
\end{algorithm}

\section{Proofs}
\label{proof:section}
\subsection{Proof of \cref{lemma:characterization}}

\subsubsection{VICReg Loss and Spectral Contrastive}
\label{proof:vic-2}

This subsection will both identify the VICReg with the spectral contrastive one through their matrix formulation.

Let us begin by reformulating the invariance term in VICReg.
For $\mZ$ defined in \eqref{eq:Z}, it is generalized to multiple pairs through
\begin{align*}
    \gL_{\rm VIC-INV} 
    &= \sum_{i,j\in[N]} \mG_{ij} \norm{\vz_i - \vz_j}^2
    = \sum_{i,j\in[N]} 2 \mG_{ij} \norm{\vz_i}^2 - 2\mG_{ij} \scap{\vz_i}{\vz_j}   
    \\& = \sum_{i,j\in[N]} 2 \mG_{ij} [\mZ \mZ^\top]_{ii} - 2\mG_{ij} [\mZ\mZ^\top]_{ji}   
    = \sum_{i\in[N]} 2[\mD \mZ \mZ^\top]_{ii} - 2[\mG \mZ\mZ^\top]_{ii}   
    \\&= 2\Tr\paren{(\mD-\mG) \mZ\mZ^\top} = 2\Tr\paren{\mZ^\top (\mD-\mG) \mZ}   
\end{align*}
where $\mD$ is the degree matrix defined as a diagonal matrix, with $A$ the number of augmented samples per original input
\[
    \mD_{ij} = \ind{i=j}\cdot \sum_{k\in[N]} \mG_{ik} = A\ind{i=j}.
\]

The variance-covariance term can be simplified by replacing the the Hinge loss for the variance by a squared norm \cite{kiani2022joint}, by setting $\beta = \alpha = 1$ and replacing $A$ by $N$ in $\mD$ to regularize diagonal terms a bit more.
Those simplifications lead to a more principled regularization term that enforces orthogonality over the dataset of the different features learned by the network $f_\theta$ \cite{cabannes2023minimal}.
The consequent regularization reads $\norm{\mZ^\top \mZ/N - \mI_K}^2$.
As a consequence, VICReg can be understood as solving for
\[
    N\gL_{\rm VIC} \approx \norm{\mZ^\top \mZ - N \mI_N}^2 + 2\trace(\mZ^\top (N\mI_N -\mG)\mZ).
\]

For the spectral contrastive Loss, it is useful to incorporate negative pairs that are sampled for the same augmentations for two different samples $\scap{\mZ_i^{(1)}}{\mZ_j^{(1)}}$ in the repulsive term.
Moreover adding $\norm{\mZ_i^{(v)}}$ on both the positive part and the negative part will not change much since $-2x + x^2$ is minimized for $x=1$.
Those modifications lead to 
\begin{align*}
    \gL_{\rm VIC^2} &= - 2\sum_{i,j\in[N]} \mG_{ij} \vz_i^\top \vz_j + \sum_{i,j\in[N]} (\vz_i^\top \vz_j)^2
    = - 2\sum_{i,j\in[N]} \mG_{ij} [\mZ\mZ^\top]_{ij} + \sum_{i,j\in[N]} [\mZ \mZ^\top]_{ij}^2
    \\&= - 2\Tr(\mG \mZ\mZ^\top) + \Tr(\mZ \mZ^\top \mZ \mZ^\top)
     = \Tr(\mZ \mZ^\top \mZ \mZ^\top - 2\mG \mZ\mZ^\top + \mG^2) - \Tr(\mG^2)
     \\&= \Tr((\mZ \mZ^\top - \mG)^2) - \Tr(\mG^2)
     = \norm{\mZ \mZ^\top - \mG}^2_F + \text{cst}.
\end{align*}
The last term being a finite constant, it can be removed from the loss.

Indeed, one can relate both the spectral contrastive loss and VICReg, by remarking that
\begin{align*}
    &\norm{\mZ^\top \mZ - N \mI_N}^2 + 2\trace(\mZ^\top (N\mI_N -\mG)\mZ)
    \\&\qquad= \trace(\mZ\mZ^\top \mZ\mZ^\top - 2N\mI_N \mZ^\top\mZ +  N^2 \mI_N)  + 2\trace(\mZ^\top (N\mI_N -\mG)\mZ)
    \\&\qquad= \trace(\mZ\mZ^\top \mZ\mZ^\top - 2\mG \mZ \mZ^\top) + N^3
    = \trace((\mZ\mZ^\top - \mG)^2) - \mG^2) + N^3
    \\&\qquad= \norm{\mZ\mZ^\top - \mG}_F^2 + \text{cst}.
\end{align*}

Finally, the variance covariance term can be written as
\begin{align*}
    \norm{\mZ\mZ^\top/ N - I}^2 
    &= \norm{\sum_{i\in[N]}\vz_i \vz_i^\top - I}^2
    = \trace\paren{\sum_{i,j\in[N]} \vz_j\vz_j^\top\vz_i \vz_i^\top - 2\sum_{i\in[N]} \vz_i\vz_i^\top + I}
    \\&= \sum_{i,j\in[N]} (\vz_j^\top\vz_i)^2 - \sum_{i,j\in[N]} (\vz_i^\top\vz_i  + \vz_j^\top\vz_j) + \text{cst}
    = \sum_{i,j\in[N]} R(\vz_i, \vz_j) + \text{cst},
\end{align*}
where $R(a, b) = (a^\top b)^\top - \norm{a}^2 - \norm{b}^2$.

\subsubsection{The SimCLR Loss}
SimCLR can be seen as a generalized linear model, where two variables $A$, $B$ are observed and the probability observing $B$ knowing $A$ is given by
\[
    p_{ij} = \Pbb(B=j\,\vert\, A=i) \propto \exp\paren{\frac{\vz_i^\top \vz_j}{\norm{\vz_i}\norm{\vz_j}}}.
\]
For simplicity, let us define $\tilde\vz = \vz / \norm{\vz}$.
SimCLR tries to maximize the likelihood of $(A, B)$ denoting random pairs coming from the same augmentations based on the observation of the graph
\[
    \prod_{ij\in[N]} p_{ij}^{\mG_{ij}} = \exp\paren{\sum_{ij\in[N]} \mG_{ij} \log\paren{\frac{\exp\paren{\tilde\vz_i^\top \tilde\vz_j}}{\sum_{k\in[N]} \exp(\tilde\vz_i^\top \tilde\vz_k)}}}.
\]
The SimCLR loss is nothing but the inverse of the log likelihood.
\[
    \gL_{\rm SimCLR} = -\sum_{ij\in[N]} \mG_{ij} \log\paren{\frac{\exp\paren{\tilde\vz_i^\top \tilde\vz_j}}{\sum_{k\in[N]} \exp(\tilde\vz_i^\top \tilde\vz_k)}}.
\]

\subsubsection{Barlow Twins}
When $\lambda = 1$, which we will consider for simplicity, the BarlowTwins loss simplifies as
\[
    \gL_{\rm BT} = \sum_{i} (1-\mC_{ii})^2 + \sum_{i\neq j} \mC_{ij}^2 = \norm{\mC - \mI_k}_F^2.
\]
Because cross-correlations are normalized cross-covariances, it is useful to introduce $\tilde\mZ$ the column normalized version of $\mZ$. Formally written in normalized matrix with the Hadamard product notation as
\[
    \tilde\mZ_{ij} = \frac{z_{ij}}{\sqrt{\sum_{k} z_{kj}^2} }
    = \frac{z_{ij}}{[(\mZ\otimes\mZ)\1]_j^{1/2}}
    \qquad\text{i.e.}\qquad
    \tilde\mZ = \mZ \diag(\mZ^{\otimes 2} \1)^{-1/2}
\]
The way the cross-correlation is built can be generalized to multiple positive pairs as
\[
    \mC_{ij} = \frac{\sum_{kl} \mG_{kl} z_{ki} z_{lj}}{\sqrt{\sum_{k} z_{ki}^2} \sqrt{\sum_{l} z_{lj}^2}}
    =\sum_{kl} \mG_{kl} \tilde{z}_{ki} \tilde{z}_{lj}
    = [\tilde{\mZ}^\top \mG \tilde\mZ]_{ij}.
\]
As a consequence, the BarlowTwins loss can be rewritten with the sole use of $\mG$ as
\[
    \gL_{\rm BT} = \norm{\tilde{\mZ}^\top\mG \mZ - \mI_K}^2_F.
\]

\subsection{The SSL Losses for Supervised Learning}

This subsection is devoted to the proof of \cref{thm:recovery}.

\subsubsection{Recovery Lemma}
\label{app:recovery}

The backbone of \cref{thm:recovery} is following Lemma.

\begin{lemma}[Equivalence between $\mY$ and $\mG$]
\label{lemma:recovery}
Given any supervised classification similarity matrix $\mG = \mY\mY^\top$ \eqref{eq:G_sup_easy}, one can recover the corresponding one-hot label encoding $\mY$, up to an orthogonal transformation $\mR$, as
\[
    \exists\mathop \mR \in O(C), \quad \text{ s.t.}\quad \mY = \mP\sqrt{\mD}\mR,
\]
where $\mP\mD\mP^T$ is the eigenvalue decomposition of the adjacency matrix $\mG(\mY)$.
Moreover the rotation $\mR$ is easily recovered by specifying the labels of $C$ samples associated with each of the $C$ different classes.
\end{lemma}
\begin{proof}
    \Cref{lemma:recovery} follows from the fact that $\mG = \mY \mY^\top$ so that $\mY$ is a square root of $\mG$, and that any two square roots of a matrix are isometric.
    In particular, if the SVD of $\mY$ is written as
    \[
        \mY = \mP \sqrt{\mD} \mR,
        \qquad \mP\in O(N), \mR \in O(C), \mD = \begin{bmatrix} \mD_1 & 0 \end{bmatrix} \in \R^{N\times C}, \mD_1 = \diag(\sigma_1^2, \cdots \sigma_C^2),
    \]
    the decomposition $\mG = PDP^\top$ is an eigenvalue decomposition of $\mG$. 
    The part $P\sqrt{D}$ is unique up to the application of a rotation on the right, which could be absorbed in $R$.

    In order to recover $\mY$ from $\mG$, notice that up to a permutation of lines and columns, $\mG$ has a block diagonal structure where each block corresponds to one label.
    If each one label is given to each block, this allows to retrieve exactly $\mY$ hence to identify $\mR$ afterwards by solving for $\mR = (\mP \sqrt{mD})^{-1} \mY$.
\end{proof}

While \cref{lemma:recovery} describes the classification case, in the generic case, if the $y$ are categorical, yet the loss $\ell(y, z)$ is not the zero-one loss, it is natural to define the similarity matrix as 
\begin{equation}
    \label{eq:gen_G_sup}
    \mG \triangleq (-\ell(y_i, y_j))_{i,j\in[N]} \in \R^{C\times C}.
\end{equation}
For example, $y_i$ could be rankings modeled with $y_i \in \Sfrak_m$ where $m! = C$, i.e. $m = \Gamma^{-1}(C) - 1$, and $\ell$ could be the Kendall loss.
In this setting,
\[
    \mG = \mY \mL\mY,
    \qquad\text{where}\qquad
    \mL \triangleq (-\ell(y, z))_{y,z\in[C]} \in \R^{C\times C},
\]
and $\mY$ is retrieved through $\mY = \mP\sqrt{\mD} \mR \mL^{1/2}$, where $\mP \mD \mP^\top$ is the eigenvalue decomposition of $\mG$, and $\mR \in \R^{C\times C}$ is an unknown rotation matrix, that might be identified by specifying at most $C$ labels associated with each of the $C$ different classes, but might be identified with a smaller number of samples if $\ell$ has a strong structure implying that $\mL$ is low-rank (see Eq. (11) in  \cite{Nowak2019}).
Indeed, the fact that compared to \eqref{eq:G_sup_easy}, the graph \eqref{eq:gen_G_sup} could be much lower rank, could lead to more efficient algorithm to image it in the active learning framework.
In essence, it would better leverage the structure encoded by the loss $\ell$.

Finally, in the regression setting, one can choose $\mG_{ij} = -y_i^\top y_j$.

\subsubsection{The VICReg Loss}
The VICReg loss is characterized as
\begin{align*}
    \gL_{\rm VIC-2} = \norm{\mZ\mZ^\top - \mG}^2_F + \text{cst}.
\end{align*}
So, it is minimized for $\mZ$ being a square root of the matrix $\mG$.
This is possible when the rank of $\mG$ which is at most $C$ since $\mG = \mY\mY^\top$ is less than the rank of $\mZ$ which is $K$.
In this setting, since $\mY$ and $\mZ$ are two square roots of $\mG$, we get
\begin{align*}
    \exists\mathop \mR \in O(C, K),\qquad \mZ = \mY \mR,
\end{align*}
where we define the rotation $O(C, K)$ as
\begin{equation}
    O(C, K) = \brace{\mR\in \R^{C\times K}\midvert \mR\mR^\top = \mI_C}.
\end{equation}

\subsubsection{The SimCLR Loss}
The probabilistic interpretation of SimCLR states that the SimCLR losses tries to maximize the likelihood of the events
\[
    \cup_{ij\in[N]} \brace{\mG_{ij}=1} \cap \brace{Y=i \,\,\&\,\, X=j},
\]
which translate as a loss in the minimization of 
\[
    \gL_{\rm SimCLR} = -\sum_{ij\in[N]} \mG_{ij} \log\paren{\frac{\exp\paren{\tilde\vz_i^\top \tilde\vz_j}}{\sum_{k\in[N]} \exp(\tilde\vz_i^\top \tilde\vz_k)}}.
\]
This is the cross entropy between $\mG_{ij}$ and $p_{ij}$ defined in the proof of the characterization of SimCLR.
If the minimization with respect to $p_{ij}$ was unconstrained, then one should match $p_{ij} \propto \mG_{ij}$.
Yet, the form of $p_{ij} \in [\exp(-1), \exp(1)]$ constraints it to go for a slightly different solution.

Remark that for two $\tilde\vz_i$, $\tilde\vz_j$ whose index $i$ and $j$ belongs to different clusters defined by the graph $\mG$, the loss is a increasing function of the quantity $\exp(\tilde\vz_i \tilde\vz_j)$.
By symmetry, we deduce that all the $\tilde\vz_i\tilde\vz_j = \cos(\vz_i, \vz_j)$ should be one for all $(i, j)$ such that $\mG_{ij}=1$.
On the other hand, the loss is a decreasing function of the $\exp(\tilde\vz_i^\top \tilde\vz_j)$ when $\mG_{ij} = 1$.
When the number of sample per class is constant, we deduce by symmetry that the different anchors for the different classes should be put at the extremity of the simplex with $C$ vertices centered at the origin and rotate with an arbitrary matrix $\mR \in O(C-1)$, which allow to recover the different classes (without their explicit labels if not provided).
When the different class have different number of samples $N_i$ with $\sum_{i\in[C]} N_i = N$, and their anchor in the output space is $\vc\in\R^K$, we are trying to minimize 
\begin{equation}
    \label{eq:anchor_simclr}
    \sum_{j\in[C]} N_j \log\paren{\sum_{i\in[C]} N_i \exp(\vc_i^\top\vc_j)},
\end{equation}
which will deform the simplex to have bigger angles between classes that are highly represented.
For example, when $N_1 = N_2 \approx N / 2$, we will have $\vc_1 \approx -\vc_2$ while the other anchors are orthogonal to one another and to $\vc_1$.
Denoting $\mM\in\R^{C}$ the matrix that maps the anchor of the class $i$ for one solution of \eqref{eq:anchor_simclr} to the $i$-th element of the canonical basis $\ve_i$ as $\vv_i \mM = \ve_i$, we get that the solution 
\[
    \mZ = \mD \mY \mR \mM^{-1}, \qquad\text{with}\qquad \mD \in \diag(\R_+^N); \mR \in O(K, C).
\]
The fact that $\mZ$ is invariant by scaling each vector $\vz$ reminds us of implementation of the cross-entropy, where to avoid divergence to infinity (since the sigmoid is optimized at infinity) one has to normalize the solution.
The SimCLR loss is actually built on the same generalized linear model as the cross-entropy, and one can roughly think of SimCLR as the SSL version of the cross-entropy.

\subsubsection{BarlowTwins}
To minimize the BarlowTwins loss
\[
    \gL_{\rm BT} = \norm{\tilde\mZ^\top \mG \tilde\mZ - I_K}_F^2,
\]
we want $\tilde\mZ$ to be a square root of the inverse of $\mG$.
To be more precise, introduce the eigenvalue decomposition of $\mG$ as $\mG = \mP \mS \mP^\top$ where $\mP \in \R^{N\times C}$ and $\mS \in \R^{C\times C}$ since $\mG$ is at most of rank $C$.
The minimizer of BarlowTwins is $\tilde\mZ = [\mP \mS^{-1/2}, 0_{N\times (K-C)}]$.
Since $\tilde\mZ$ is the column normalized version of $\mZ$, $\mZ$ can be reconstructed for any diagonal matrix $\mD \in \diag(\R_+^{K})$ as $\mZ = \tilde\mZ \mD$.
Incorporating $[\mS^{-1}, 0]$ in $\mD$, we get that the minimizer of the BarlowTwins loss are exactly the matrices $\mZ = \mP \mS^{1/2}[\mD, 0_{K-C}]$ for $\mD \in \diag{\R_+^C}$.
Moreover, since both $\mP \mS^{1/2}$ and $\mY$ are square root of $\mG$, we know that the exists a rotation matrix $\mR \in O(K)$ such that $\mP \mS^{1/2} = \mY \mR$. 
All together, we get that the minimizer of the BarlowTwins loss are exactly the
\[
    \mZ = \mY\mR\mD, \qquad\text{for}\quad \mR \in O^{C, K},\, \mD \in \diag(\R_+^C)
\]
The fact that BarlowTwins do not care about the amplitude of the solution $\mZ$ reminds us of discriminant analysis that learns classifiers by optimizing ratio and angles.

\subsection{Bayes optimum}
\label{app:bayes}

For completeness, we now state a Bayes optimum proposition regarding the VICReg loss of the paper.

\begin{proposition}[Bayes optimum]
    \label{prop:bayes}
    When $K \geq C$ and there is no context, i.e. $\vx_i = \vx_0$ for all $i\in [N]$, and $y_i$ are sampled according to a noisy distribution $\paren{y\midvert \vx=\vx_0}$, the naive study of the VICReg Bayes optimum is meaningless, since
    \begin{align*}
        \argmin_{\vz \in \R^K} \gL_{\rm VIC-2}(\begin{bmatrix} \vz\\ \vdots \\ \vz \end{bmatrix};\mG) =  \brace{\vz\in\R^K\midvert \norm{\vz} = 1}.
    \end{align*}
    Yet, if one free the variable $\mZ \in \R^{N\times K}$, we have
    \begin{align*}
        \argmin_{\mZ \in \R^{N\times K}} \gL_{\rm VIC-2}(\mZ;\mG) =  N^{1/2}\cdot\brace{(\Pbb(Y=i)^{1/2} e_i)_{i\in[C]}\cdot \mR \midvert \mR \in O(C, K)},
    \end{align*}
    where $(e_i)_{i\in[K]}$ is the canonical basis of $\R^K$.
\end{proposition}

\begin{proof}
    For the first part of the proof, remark that the invariance term in VICReg will be zero for any $\vz$, so VICReg loss is minimized for any vector that minimized the variance-covariance term $\norm{\vz\vz^\top - I}^2$, which is done for any unit vector.
    
    For the second part, remark that $\mG = \mY\mY^\top$ has $C$ connected components, that are all full cliques, i.e. the adjacency is filled with one.
    As a consequence, the eigenvectors of $\mG$ associated with non-zeros elements are exactly the $(\ind_{y_i = y})_{i\in[N]}$ for $y\in [C]$, and the corresponding eigenvalues are $N_y$ where $N_y = \sum_{i\in[N]} \ind{y_i = y} = N\Pbb(Y=i)$ are the number of element in the class $i\in[C]$.
    As a consequence, a square root of $\mG$ is $N^{1/2} (\Pbb(Y=i)^{1/2} \delta_{ij})_{i\in[C], j\in[K]}$, hence the proposition following the fact that all the square root of $\mG$ are isomorphic.
\end{proof}

\section{Additional experimental details}
\label{app:expe}

\subsection{Essential Code}
\label{sec:code}

{\bf SSL Graph}~
\begin{lstlisting}[language=Python,escapechar=\%]
G = torch.zeros(N * V, N * V)                                   #  X in R^{Np x D}, V views
i = torch.arange(0, N * V).repeat_interleave(V - 1)             # row indices
j = (i + torch.arange(1, V).repeat(N * V) * N).remainder(p * V) # column indices
G[i,j] = 1                                                      # unweighted graph
\end{lstlisting}

\medskip
{\bf Sup Graph}~
\begin{lstlisting}[language=Python,escapechar=\%]
Y = torch.nn.functional.one_hot(labels, num_classes=num_classes).float()
G = Y @ Y.T
\end{lstlisting}

\medskip
{\bf VICReg.}~
\begin{lstlisting}[language=Python,escapechar=\%]
C = torch.cov(Z.t())                                     # Z in R^{N x K}
reg_loss = torch.nn.functional.mse_loss(C, torch.eye(K)) 
reg_loss *= out_dim ** 2                                 # correct for mean vs sum
i,j = G.nonzero(as_tuple=True)
inv_loss = torch.nn.functional.mse_loss(Z[i], Z[j])      # pairwise L2 weighted by G_{i,j}
inv_loss *= out_dim
loss = beta * inv_loss + reg_loss
\end{lstlisting}

\medskip
{\bf SimCLR}~
\begin{lstlisting}[language=Python,escapechar=\%]
Z_renorm = torch.nn.functional.normalize(Z, dim=1)          # Z \in \mathbb{R}^{N \times K}
cosim = Z_renorm @ Z_renorm.t() / tau                       # N x N matrix, tau is the temperature 
mask = 1 - torch.eye(N, N, device=Z.device, dtype=Z.dtype)
loss = (G * (torch.logsumexp(cosim*mask, dim=1, keepdim=True) - cosim)).mean()
\end{lstlisting}

\medskip
{\bf SCL}~
\begin{lstlisting}[language=Python,escapechar=\%]
Z = torch.nn.functional.normalize(Z, dim=1)
loss = torch.nn.functional.mse_loss(G, Z@Z.T)
\end{lstlisting}

\subsection{Controlled experiments}
\subsubsection{Setup}

\begin{figure}[ht]
    \centering
    \includegraphics{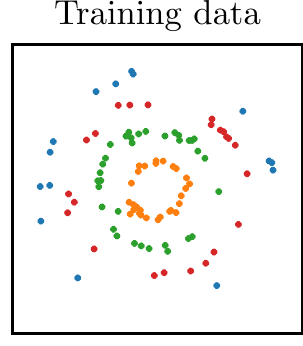}
    \includegraphics{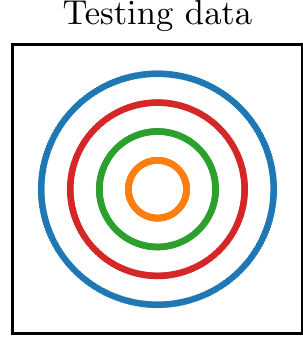}
    \caption{Setup for the controlled experiments of \cref{fig:active}.
    The dataset is made of four concentric circles that corresponds to four different classes represented by different colors. The training dataset is made of one hundred random points, with some noise.}
    \label{fig:setup_toy}
\end{figure}

\begin{figure}[ht]
    \centering
    \includegraphics[width=.2\linewidth]{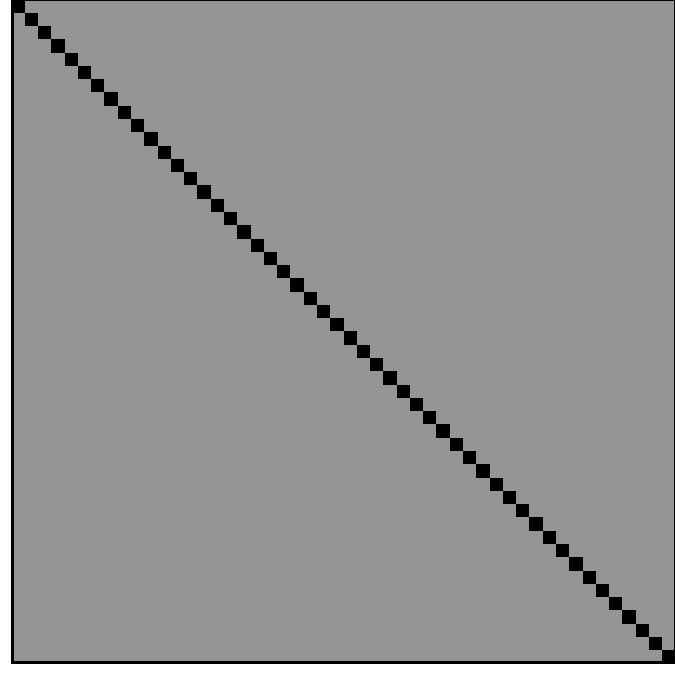}
    \includegraphics[width=.2\linewidth]{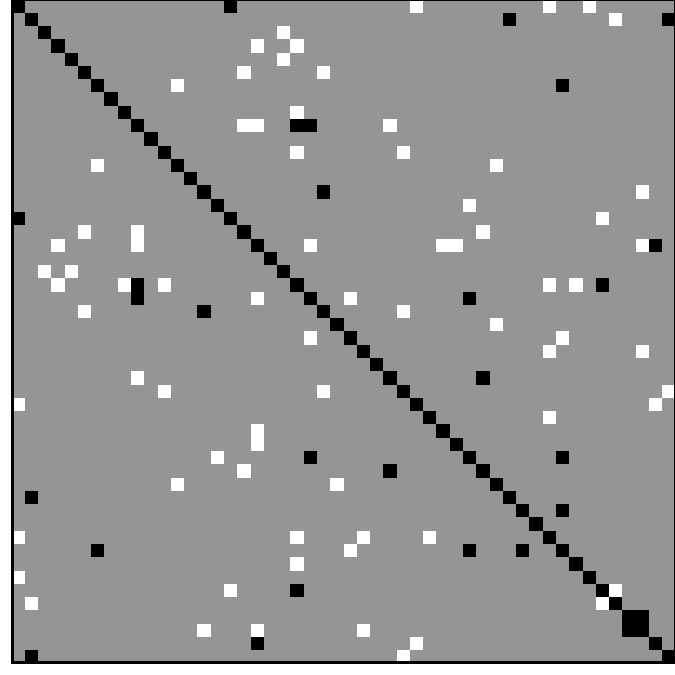}
    \includegraphics[width=.2\linewidth]{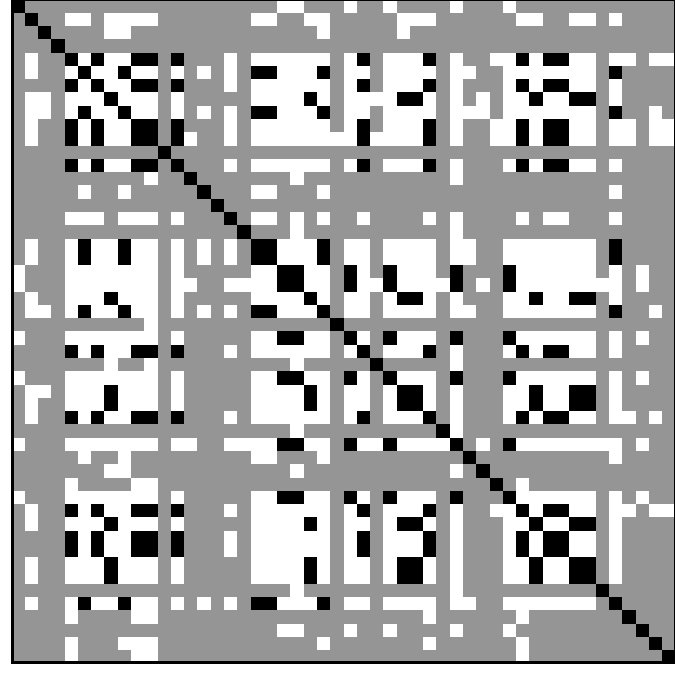}
    \includegraphics[width=.2\linewidth]{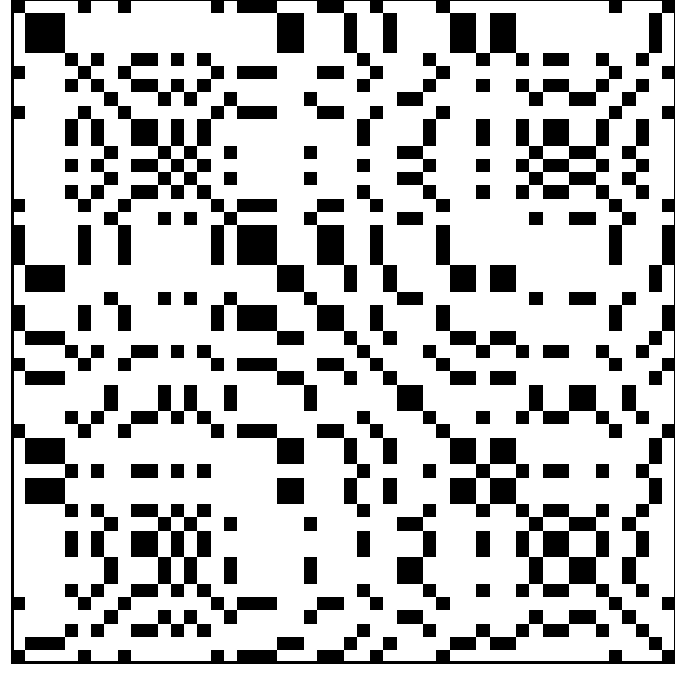}
    \caption{
    Graphs $\mG$ corresponding to the different snapshot taken on \cref{fig:active}.
    Grey indicates zeros, white indicates negative observations, and black means positive ones.
    The main strength of the active strategy in \cref{alg:active} is that, by leveraging the underlying structure of the graph, is able to deduce much faster the full graph $\mG$ than the naive passive implementation that only asks for random query pairs.
    Basically a positive observation is turned into many negative observations.
    }
    \label{fig:graph_fig4}
\end{figure}

The train and test set of Figure \ref{fig:active} is shown on \cref{fig:setup_toy}.
The similarity graphs corresponding to the different snapshots on \cref{fig:active} are shown on \cref{fig:graph_fig4}.
In all the experiments, we consider $K = C + 1 = 5$.

\subsubsection{Contrastive vs. Non-Contrastive}

\begin{figure}[ht]
    \centering
    \includegraphics{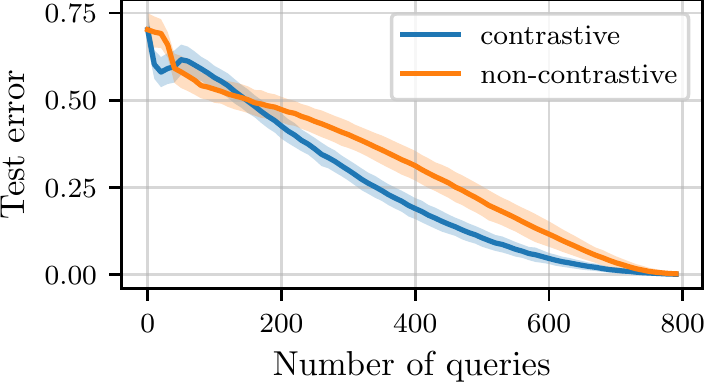}
    \caption{Comparison of contrastive ($\mG_{ij} \in \brace{-1, 0, 1}$) and non-contrastive ($\mG_{ij} \in \brace{0, 1}$) variation of VICReg with $N=300$.
    The setting is the same as \cref{fig:active} with \cref{alg:active}.
    We remark the usefulness to distinguish between negative pairs and unknown pairs, although some instability issues seem to appear when few entries are known for the contrastive method.
    }
    \label{fig:contrastive}
\end{figure}

Intuitively, it is useful to distinguish more explicitly between positive, negative and unknown relations, which we test on \cref{fig:contrastive}.
To do so, the graph $\mG$ is modified to encode semantically similar elements as positive edges, dissimilar ones as negative edges, while unknown relationships are going to be represented by zeros.
\begin{equation}
    \mG_{ij} = \left\{\begin{array}{cl} 1 & \text{if } \vx_i\sim \vx_j \text{ has been observed}, \\ -1 & \text{if } \vx_i \not\sim \vx_j \text{ has been observed}, \\ 0 & \text{otherwise}. \end{array} \right.
\end{equation}
One might wonder if this really improves performance. 
The comparison is the object of \cref{fig:contrastive}

\subsubsection{The Benefits of Incorporating Known Labels}
A major motivation of this paper is to be able to add prior information on sample relationships in SSL methods, and more in particular, to have a simple way to leverage known labels.
Let us denote by $\hat\mY \in\R^{N\times D}$ the one-hot matrix $(\vy_i)_{i\in[N]}$ where $\vy_i$ is the one-hot vector of the label $y_i$, such that if $y_i$ is unknown $\vy_i = 0$.
The knowledge of some coefficients of the real $\mY$, leads to the knowledge of a few coefficient of $\mG^{(\sup)} = \mY\mY^\top$, those could be added to the SSL graph to add useful connection deduced from the labels, leading to
\[
    \mG = (1-\alpha) \cdot \mG^{(\ssl)} + \alpha \cdot \hat\mY\hat\mY^\top,
\]
where $\alpha \in [0, 1]$ is a mixing coefficient stating how much the supervised information should weigh in the similarity matrix.
Naively, we could set $\alpha=1/2$, yet when only few labels are given this would destabilize the spectral decomposition of $\mG$ too much, and we observe on Figure \ref{fig:interpolation} that a small mixing coefficient is better.
An explanation could be that the relations encoded by SSL are quite local and subtle, while the connections suggested by supervised learning are quite global and brutal on it suggested to fold the input space, hence need to be dampened when mixing the SSL and supervised graphs.

\subsubsection{Robustness to noise}

\begin{figure}
    \centering
    \includegraphics{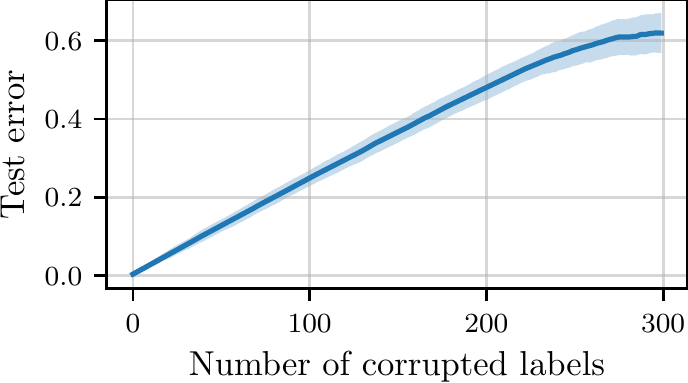}
    \caption{Study of the effect of labeling noise.
    The setup is the same as \cref{fig:active} yet with $N=300$ points.
    We consider having full access to $\mY$ thus to $\mG = \mY\mY^\top$ yet we assume that a certain number of labels $y_i$ are corrupted. 
    We see that the algorithm is somewhat robust to noise.}
    \label{fig:noise}
\end{figure}

As mentioned in the last part of the paper, depending on the algorithm used, the effect of noise in queries answers might lead to dramatic performance loss.
In the main text, we were careful to describe algorithms that are robust to noise.
The effect of noise in the labels for \cref{alg:active} is studied in \cref{fig:noise}.
Because of its structure, noise in the query for \cref{alg:active} is equivalent to noise in the label $\vy$.
This explains the setup of the figure.

\subsubsection{The Importance to Recover Connected Components}

\begin{figure}[ht]
    \centering
    \includegraphics{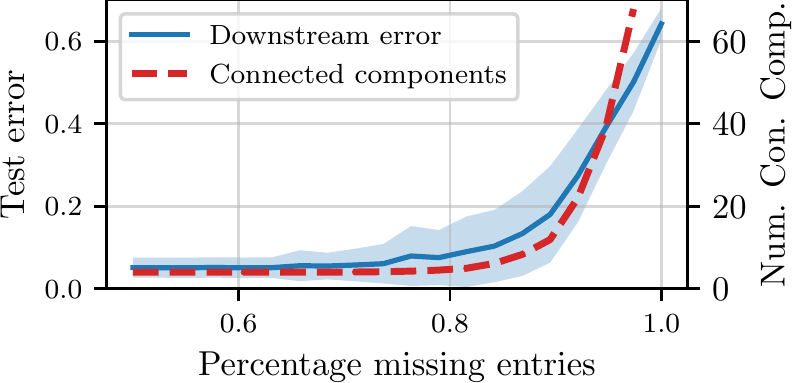}
    \caption{Comparison between the test error and number of connected components in the graph $\mG$ as a function of the percentage of missing entries in $\mG$.
    The test error is reported as in \cref{fig:active}, but it is reported as a function of missing entries of the supervised learning graph $\mG^{(\sup)}$.
    The standard deviation for the red curve is not represented here as the number of connected components is highly concentrated around its mean.
    }
    \label{fig:comp}
\end{figure}

An interesting experiments is provided by \cref{fig:comp}, which compares the test error and the number of connected components of the graph $\mG$ as a function of the number of missing entries of $\mG = \mG^{(\sup)}$. 
In our synthetic experiment, $\mG^{(\sup)}$ has four connected components corresponding to the four classes in the dataset, e.g. \cref{fig:G}.
Typically, based on transitivity of the similarity relation $\sim$, one can hope to only need $O(1/N) = O(NC / N^2)$ queries, i.e. reconstructed entries of $\mG$, to have a good sense of the global $\mG$, hence to learn $f_\theta$.
Moreover, on \cref{fig:comp}, the test error can be relatively well-predicted by the number of connected components of the graph $\mG$.
This suggests creative ways to design active learning strategies based on search to optimize the number of connected components of $\mG$.
However, leveraging transitivity of the similarity relationship to fill $\mG$ efficiently might be limited when queries answers are noisy, although literature on error correcting codes might be useful \cite{Varshamov1957,Cover1991}.
Moreover, the binary (and transitive) nature of similarity can be questioned when SSL sometimes uses DA that provides iconoclast unrealistic images, and one might prefer to assign similarity scores.
Problems that do not occur with the transitivity agnostic \cref{alg:active}.

\subsubsection{Mixture of Gaussian}

\begin{figure}
    \centering
    \includegraphics{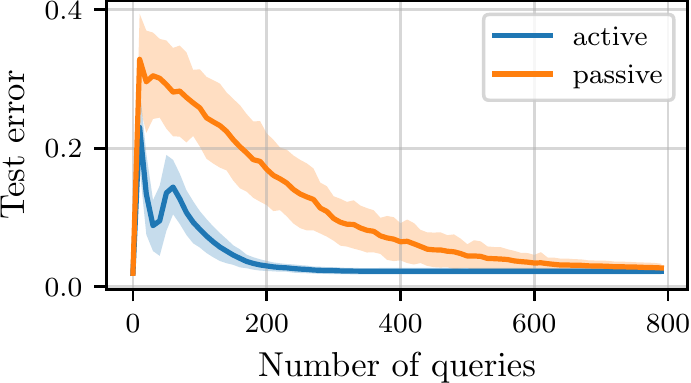}
    \hspace{1em}
    \includegraphics{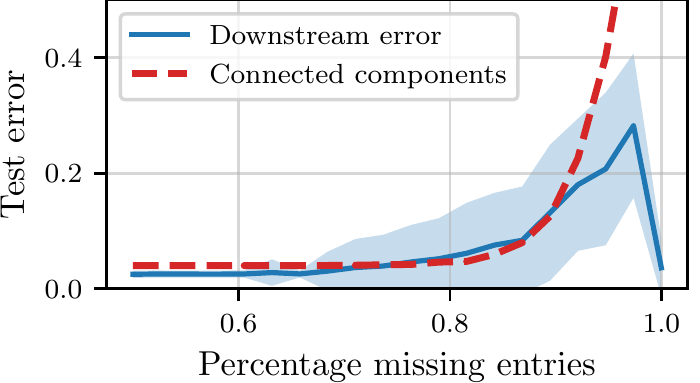}
    
    \vspace{1em}
    \includegraphics{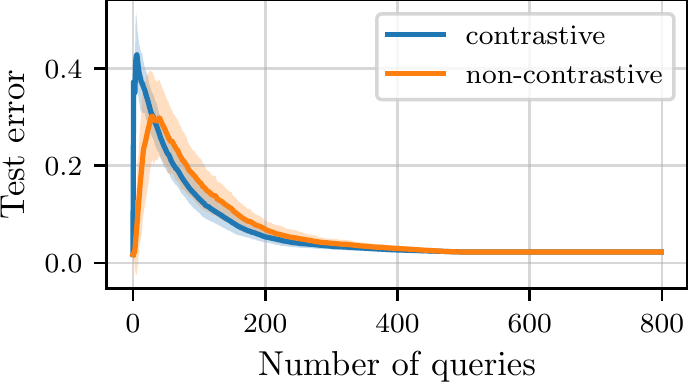}
    \hspace{1em}
    \includegraphics{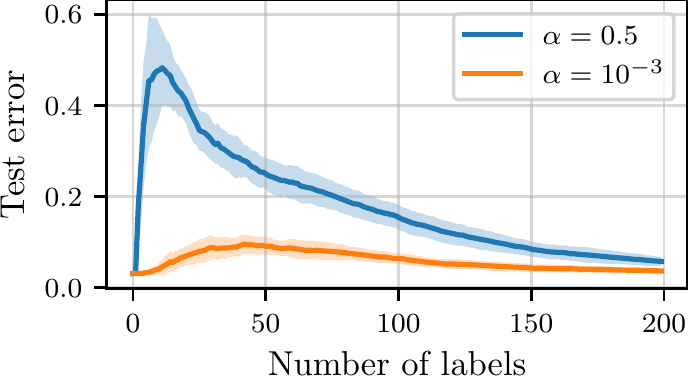}
    
    \vspace{1em}
    \includegraphics{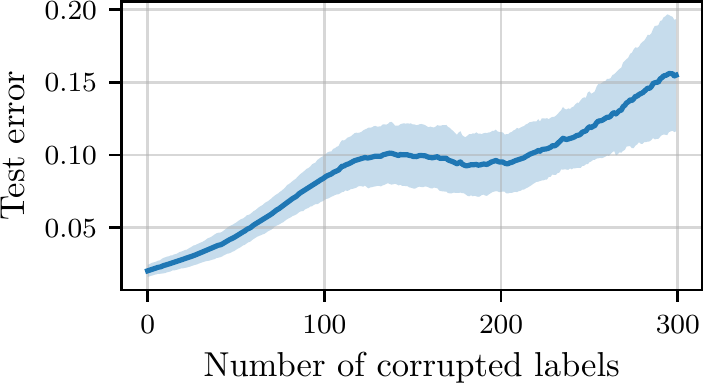}
    \caption{Same figures as before with a mixture of Gaussians dataset.
    The mixture dataset has the particularity that the downstream task can be solved with any orthogonal basis of $\R^C$. 
    When no queries has been made, $\mG = I_N$, and the spectral decomposition of this graph will lead to a representation that can solve the downstream task, explaining why when no queries have been made, or when all the entries of $\mG$ are removed, the downstream task can be solved.}
    \label{fig:gaussian}
\end{figure}

One can question if the findings presented so far are specific to the concentric circles datasets.
In order to assert the validity of those findings, we consider a second dataset, made of mixture of Gaussian, formally
\[
    \mX = \mY + \sigma\mE, \qquad\text{where}\qquad
    \mE_{ij} \sim {\cal N}(0, 1),
\]
given a label $y\in[C]$, $\vx$ is generated according to ${\cal N}(\ve_y, \sigma I_C)$.
The results are reported on \cref{fig:gaussian} with $\sigma=.3$.

\subsection{Real-world experiment}

While it is hard to control all the factors that come into play when training a neural network on real data, our experiments suggested that what we have seen in controlled experiments transfer to real-world problems.
In particular, we consider the CIFAR-10 dataset, with a resnet 18 architecture.
A first stage was representation learning, where we used the VICReg loss to learn representation with the CIFAR-10 training set.
In particular, we removed the classifier head of the resnet and replaced it with two fully connected layers with batch norms.
The number of output dimensions was set to $K = 16$, and the number of hidden neurons was set to $4K$.
After the representation was learned, we replaced the classifier head by a linear layer with $K=C$ output dimension and fit this last layer on the CIFAR-10 training set.
The resulting network was then tested on the CIFAR-10 testing set.
Regarding hyperparameters (network, DA, optimizer), we fixed them in accordance with tutorial online (in particular the pytorch-lightning tutorial) in order to achieve high performance results on CIFAR with SSL.
In our first experiments, we stopped after two epochs of training for pretraining (since the output dimension is quite small, there is no need to go really far away in training), and twenty epochs downstream.
The pretraining task consisted in all the training data of CIFAR-10 tackle, 
We found that the representation learned with SSL was achieving 28 \% accuracy on CIFAR-10 with linear probing, while the representation learned directly with the supervised graph was achieving 63 \% accuracy.
In the meanwhile, training a resnet with classifier head to be made of 60 hidden neurons and 10 output dimensions with the ground truth labels and the mean-square error in the exact same setting leads to a performance of 63\% too.
In other terms, in these simple experiments, one can use the VICReg technique we derived here, or the MSE loss and get the same performance.
Training for tens epochs for the upstream task (the minimization of the VICReg loss), and one hundred for the downstream one (the linear head fitting), we improved performance to 62\% for SSL and 66\% for the supervised learning graph.
Furthermore, we did not perform extensive hyperparameter tuning, which suggests that the supervised learning performance could be even more competitive, since we took parameters that are known to be good for the self-supervised learning techniques.
All the code is available to reproduce our experiments.

\end{document}